\documentclass[acmtog]{acmart}

\pdfoutput=1

\usepackage[utf8]{inputenc} % allow utf-8 input
\usepackage[T1]{fontenc}    % use 8-bit T1 fonts
\usepackage{hyperref}       % hyperlinks
\usepackage{url}            % simple URL typesetting
\usepackage{booktabs}       % professional-quality tables
\usepackage{amsfonts}       % blackboard math symbols
\usepackage{amsmath}       % blackboard math symbols
\usepackage{nicefrac}       % compact symbols for 1/2, etc.
\usepackage{microtype}      % microtypography
\usepackage{xcolor}         % colors
\usepackage{soul}
\usepackage[normalem]{ulem}

\usepackage{color}
\usepackage{graphicx}
\usepackage{layouts}

\setcitestyle{square} % was nosort, square ;nosort to allow for manual chronological ordering

\definecolor{turquoise}{cmyk}{0.65,0,0.1,0.3}
\definecolor{purple}{rgb}{0.65,0,0.65}
\definecolor{dark_green}{rgb}{0, 0.5, 0}
\definecolor{orange}{rgb}{0.8, 0.6, 0.2}
\definecolor{red}{rgb}{0.8, 0.2, 0.2}
\definecolor{darkred}{rgb}{0.6, 0.1, 0.05}
\definecolor{blueish}{rgb}{0.0, 0.3, .6}
\definecolor{light_gray}{rgb}{0.7, 0.7, .7}
\definecolor{pink}{rgb}{1, 0, 1}
\definecolor{greyblue}{rgb}{0.25, 0.25, 1}
\definecolor{tab_blue}{HTML}{1f77b4}
\definecolor{tab_orange}{HTML}{ff7f0e}
\definecolor{LightRed}{rgb}{0.99,0.89,0.89}
\definecolor{mesh_misty_rose}{HTML}{e6aaa3}
\definecolor{mesh_yellow}{HTML}{ffba00}
\definecolor{MyDarkBlue}{rgb}{0.02,0.02,0.6}

\newcommand{\todo}[1]{}%{\color{red}{\bf [TODO: #1]}}}
\newcommand{\now}[1]{}%{\color{purple}{\bf [TODAY: #1]}}}
\newcommand{\green}[1]{}%{\color{dark_green}{#1}}}
\newcommand{\PS}[1]{}%{\color{darkred}{\bf [PS: #1]}}}
\newcommand{\VD}[1]{}%{\color{tab_blue}{\bf [VD: #1]}}}
\newcommand{\JP}[1]{}%{\color{dark_green}{\bf [JP: #1]}}}
\newcommand{\MG}[1]{}%{\color{teal}{{\bf MG:} #1}}}
\newcommand{\RM}[1]{}%{\color{red}{\sout{#1}}}}
\newcommand{\fredo}[1]{}%{\color{purple}{\bf [Fredo: #1]}}}
\newcommand{\bill}[1]{}%{\color{turquoise}{\bf [Bill: #1]}}}
\newcommand{\ct}[1]{#1} %check true
\newcommand{\new}[1]{#1}
\newcommand{\update}[1]{#1}

\newcommand{\crosssim}{Cross-Similarity Feature Weighting}

\title{Materialistic: Selecting Similar Materials in Images}
\author{Prafull Sharma}
\affiliation{%
	\institution{MIT}
	\country{USA}
}
\affiliation{%
	\institution{Adobe Research}
	\country{USA}
}
\email{prafull@mit.edu}

\author{Julien Philip}
\affiliation{%
	\institution{Adobe Research}
	\country{UK}
}
\email{juphilip@adobe.com}

\author{Michael Gharbi}
\affiliation{%
	\institution{Adobe Research}
    \country{US}
}
\email{mgharbi@adobe.com}

\author{Bill Freeman}
\affiliation{%
	\institution{MIT}
     \country{US}
}
\email{billf@mit.edu}

\author{Fredo Durand}
\affiliation{%
	\institution{MIT}
     \country{US}
}
\email{fredo@mit.edu}

\author{Valentin Deschaintre}
\affiliation{%
	\institution{Adobe Research}
     \country{UK}
}
\email{deschain@adobe.com}

\setcopyright{acmlicensed}
\acmJournal{TOG}
\acmYear{2023} 
\acmVolume{42} 
\acmNumber{4} 
\acmArticle{1} 
\acmMonth{8} 
\acmPrice{15.00} 
\acmDOI{10.1145/3592390}

\begin{document}
\begin{abstract}
    Separating an image into meaningful underlying components is a crucial first step for both editing and understanding images.
    We present a method capable of selecting the regions of a photograph exhibiting the same material as an artist-chosen area.
    Our proposed approach is robust to shading, specular highlights, and cast shadows, enabling selection in real images.
    As we do not rely on semantic segmentation (different woods or metal should not be selected together), we formulate the problem as a similarity-based grouping problem based on a user-provided image location.
    In particular, we propose to leverage the unsupervised DINO~\cite{caron2021emerging} features coupled with a proposed \crosssim{} module and an MLP head to extract material similarities in an image.
    We train our model on a new synthetic image dataset, that we release.
    We show that our method generalizes well to real-world images.
    We carefully analyze our model's behavior on varying material properties and lighting. Additionally, we evaluate it against a hand-annotated benchmark of 50 real photographs.
    We further demonstrate our model on a set of applications, including material editing, in-video selection, and retrieval of object photographs with similar materials. Project website: \href{https://prafullsharma.net/materialistic/}{https://prafullsharma.net/materialistic/}
\end{abstract}



\begin{teaserfigure}
\includegraphics[width=\textwidth]{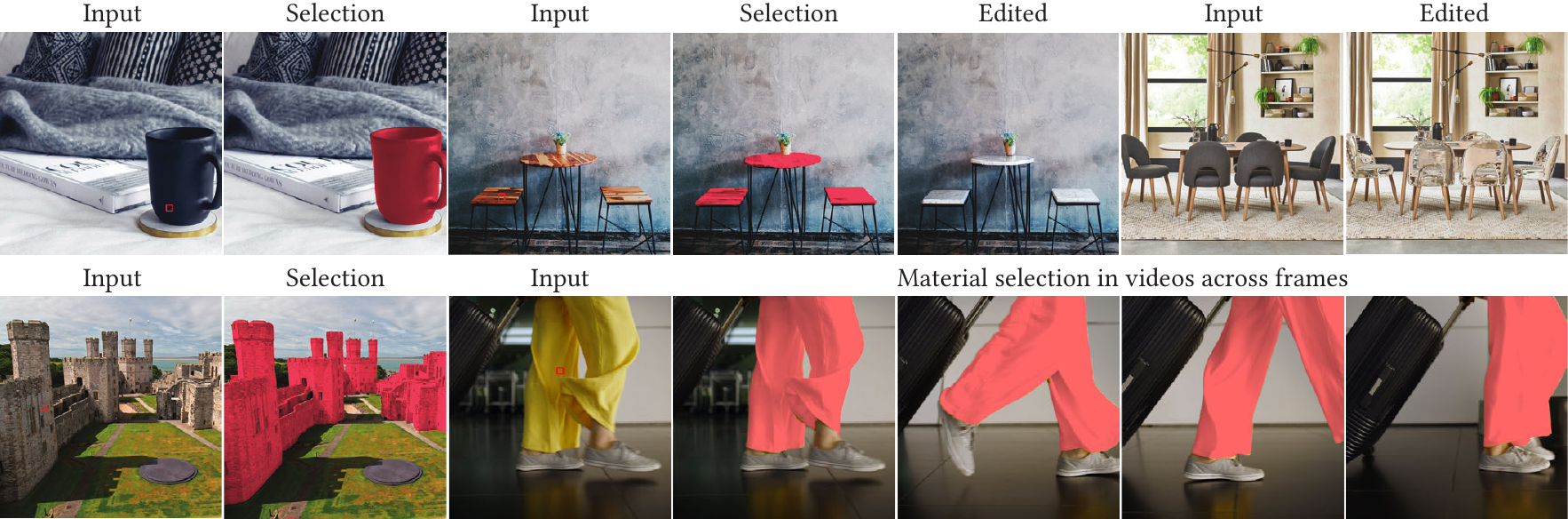}
\caption{Given an input image and a pixel selection (marked with a red square), our method automatically selects all the pixels that have the same material as the query.
Our algorithm can identify materials shared by different objects (table and stools), and is robust to shading variations (castle example).
Material-based selection enables downstream applications, such as material editing or replacement (top right).
% demonstrating its ability of identifying the materials in an image.
%
Our approach can be extended to select materials across multiple images, enabling material selection in videos (bottom right) \emph{without} requiring any optical flow propagation.}
\label{fig:teaser}
\end{teaserfigure}

\maketitle

\section{Introduction}
In this work, we present a method to select image regions with the same material as a given query pixel.
This enables a wide variety of image editings as shown in Figure~\ref{fig:teaser}, and can be used to guide down stream-tasks such as inverse rendering~\cite{nimierdavid2021material}.
Material selection is a challenging ill-posed problem because a material's appearance can vary drastically within a single image, depending on the viewing angle and the local illumination, such that two pixels with the same material can exhibit very different reflected colors and intensities.
Since a pixel's color is a complex function of the scene's geometry, illumination, and materials, the converse can also hold: two pixels with the same radiance, may belong to different materials.
Yet, humans can identify objects that share the same material with surprising accuracy, regardless of an object's shape, and despite shading variations and strong light-dependent effects, such as specular reflections and cast shadows.
It is remarkable, for instance, that we can, from a single image, identify that both chairs and the table in Fig~\ref{fig:teaser} are made of the same wooden material.
In contrast to semantic segmentation, our method does not rely on a predetermined closed set of material classes.
Instead, it dynamically evaluates material similarities between a user query location and all other pixels.
This approach generalises to materials which have not been seen during training.
We further show our method is robust to variations in shading, and in the geometries on which the material appears.

In this paper, we consider that two surfaces have the same material if they share the same texture and reflectance properties.
For instance, we consider a wood with growth color variations, or a wallpaper with small repeating patterns to be single materials.
However, we consider two woods with different grain textures, or different colours to be distinct materials.
To the best of our knowledge, ours is the first method that can select image regions based on the material at a user-selected pixel. 
Existing selection tools either perform selections based on color or intensities, requiring continued user interactions (e.g., the ``lasso'' tool), or perform object-level selections, using semantic and instance-level segmentation models~\cite{he2017mask,cao2020d2det,yuan2018ocnet,tan2019efficientnet,wang2020axial,wang2020solov2}.
Color-based methods and texture segmentation methods~\cite{todorovic2009texel,belongie1998color,deng2001unsupervised,haindl2008texture, knn_matting} are not robust to shading variations, as shown in Figure~\ref{fig:comparison}.
Existing material segmentation approaches~\cite{upchurch2022dense, bell15minc} do not provide sufficient granularity: they are limited to a fixed set of high-level, predefined material classes (e.g., wood vs.\ metal). 
This precludes selecting a specific wood material in a scene containing distinct types of wood for example.
Object selection methods built on instance-level segmentation are closer in spirit to our approach, but cannot perform precise material selections since a single object can be made of several materials, and a given material can appear on multiple objects.

To perform in-the-wild natural image material selection, we use a pre-trained self-supervised vision transformer, DINO~\cite{caron2021emerging}, as a fixed feature extractor to compute a patch-level representation of the input image, leveraging its natural image priors.
We then specialize these generic pretrained features for material selection using a multi-scale neural network.
To specify the query pixel, we propose a \crosssim{} mechanism that modulates features at different resolutions and fuses them to obtain a material similarity score.
We train our model exclusively on a synthetic dataset containing 50,000 images of indoor scenes rendered using a physically-based path tracer. The images were rendered using 100 indoor scenes with defined camera trajectories and 16,000 physically-based rendering materials.

Despite being trained on synthetic indoor scenes, we show that our model exhibits great generalization to real photographs, including outdoor images. 
Further, our method supports cross-image selections: a query embedding from a given image can be used to select similar materials in other images. 
This enables material selection in videos or material-based image retrieval in object picture databases.
We demonstrate these applications and analyze the behavior of the method through a set of controlled experiments that vary material, color, lighting, selection position, and image resolution.

In summary, we propose a method that adds to the palette of image selection tools, simplifies a wide range of editing tasks, and provides important information for downstream tasks like material recognition and acquisition.
We enable this through the following key contributions:
\begin{itemize}
    \item The first material selection method for natural images, robust to shading and geometric variations.
    \item A novel, query-based, architecture inspired by vision transformers, allowing to select pixels based on user input.
    \item A new large dataset of photorealistic synthetic HDR images with per-pixel fine-grained material labels.
\end{itemize}

\section{Related Work}
\label{sec:RW}

\paragraph{Semantic segmentation}
Parsing a scene into "things and stuff" \cite{adelson2001seeing} is critical for scene understanding.
It has led to the development of semantic and instance segmentation datasets with per-pixel annotations of object classes~\cite{zhou2017scene,Everingham15,lin2014microsoft,cordts2016cityscapes,geiger2012we,caesar2018coco}.
Fully convolutional networks~\cite{long2015fully} have become the standard for image segmentation, with ever-improving efficiency and accuracy on established benchmarks~\cite{tan2019efficientnet,cao2020d2det,chen2017deeplab,wang2020solov2,he2017mask}.
Recent methods have also explored open vocabulary image segmentation~\cite{ghiasi2021open}.
Although critical steps towards scene understanding, these methods are often limited by to the fixed set of labels or vocabulary they use during training
and cannot adjust their segmentation for previously unseen labels.
In contrast, our method can dynamically adapt its selection to a user query.

\paragraph{Material classification}
Material classificaton is a long standing problem~\cite{leung2001representing}; it aims at recognising the type of material in an image based on a pre-defined set of classes. Prior to deep learning, methods relied on various filter banks~\cite{leung2001representing,fogel1989gabor} to extract relevant features for classification. Based on improvements in deep learning for image segmentation, per-pixel classification architecture were proposed for material types (metal, wood, \ldots)~\cite{Schwartz_2013_ICCV_Workshops, cimpoi2014deep, bell15minc, 
Schwartz2016MaterialRF} and material properties, such as fuzziness~\cite{Schwartz20}. Specialized angular imaging capture systems were also proposed to further improve automatic material classification in the wild~\cite{Xue2022}.
Unlike ours, these approaches are limited to a pre-defined set of material classes, and therefore cannot handle materials outside their label set.
Further, they group different variations of a material in the same generic class (e.g. `wood'), despite strong intra-class appearance variations.

\paragraph{Material segmentation}
Prior work has also extensively studied texture segmentation using co-occurrence matrices~\cite{haralick1973textural}, EM~\cite{belongie1998color}, filtering~\cite{randen1999filtering,reyes2006bhattacharyya}, and watershed~\cite{malpica2003multichannel}.
These methods can segment contiguous texture regions, but do not handle disjoint regions, e.g., when multiple objects have the same material,
and they do not enable a user input to specify the selection.
Flat surface material segmentation typically relies on Matrix Factorization~\cite{Lawrence2006} or scribble interfaces, letting user control the segmentation ~\cite{Pellacini07, Lepage11, knn_matting, hu2022}. These approach are however limited to the BTF/SVBRDF domain and cannot handle natural images.
Different methods were proposed for scribble-based natural image decomposition into intrinsic images~\cite{Bousseau2009}, or for color and local statistics based image editing~\cite{An08}.
Their assumptions are too limiting for our material selection task.

\paragraph{Attention models and Vision Transformers}

Recently, the attention mechanism~\cite{vaswani2017attention} has been used in several vision tasks, such as image classification~\cite{dosovitskiy2020image,chen2020nonlocal,hu2018squeeze}, semantic segmentation~\cite{DPT,wang2021max,wang2020axial}, super-resolution~\cite{yang2020learning,cao2021video,lu2021efficient,chen2022activating}, image generation~\cite{zhang2022styleswin,peebles2022scalable}, and self-supervised visual representation learning~\cite{deng2001unsupervised,caron2021emerging}.
Specifically, DINO~\cite{caron2021emerging} uses a Vision Transformer (ViT) to performs self-distillation to learn visual representations and demonstrate unsupervised class-specific salient object segmentations. Moreover, STEGO~\cite{hamilton2022unsupervised} uses DINO as the backbone visual representation to extract dense semantic correspondence between images and semantic segmentations. %
Likewise, we build our material selection model atop pre-trained DINO features, benefiting from their large-scale real image prior.
Also, our \crosssim{} layer injects the user's pixel query into our model using a cross-similarity weighting scheme inspired by cross-attention~\cite{vaswani2017attention}. 

\paragraph{Material inverse rendering}
Inverse rendering for materials aims at recovering appearance properties of materials from image(s)~\cite{Guarnera2016}.
This problem is inherently ill-posed. 
So, several methods rely on data-learned prior~\cite{Deschaintre2018, Deschaintre2019, Deschaintre2021, Li2018, li2018_3D, Gao2019, li2020inverse, Guo2020, Guo2021, zhou2022tilegen}.
Other methods rely on a stationarity assumption ~\cite{Aittala2015, Aittala2016, Deschaintre2020, Henzler2021} to extract material information from a few images. 
Inverse rendering methods~\cite{azinovic2019inverse, nimierdavid2021material} often explicitly require material segmentation as an input.
\citet{hu2022control} use it to improve their material editing results.
Our material selection approach is orthogonal to these works. It can be used as guidance to these methods, to better share information across the image and facilitate inverse rendering.

\paragraph{Material datasets}

Multiple datasets providing some level of material information have been proposed for classification, segmentation~\cite{liu2010exploring,schwartz2019recognizing,bell2013opensurfaces, upchurch2022dense,bell15minc}, material inverse rendering~\cite{Deschaintre2018} or image editing tasks such as relighting~\cite{Philip19, Philip21, griffiths2022outcast, murmann2019dataset, 10.1145/3384382.3384523}.
However, these datasets do not contain fine per-pixel material annotations.
Some of them do contain per-pixel material \emph{class} information (\textit{i.e.} wood, metal), but they do not differentiate between intra-class instances.
The lack of fine-grained material annotation creates false positives for our task (two different woods would be in the same class) and prevents us from using these datasets.
As a result, we chose to render a new synthetic dataset containing $50,000$ images of indoor scenes and per-pixel material segmentation which respects the intra-class variations of materials. 

\paragraph{Selection tools} 
Typical selection tools (e.g., those found in Photoshop) include the lasso or "magic wand", based on color information, as well as object-based selection tools, built on semantic and instance segmentation technologies.
Color-based selection tools are insufficient for material selection, because lighting and geometric variations can significantly alter the reflectance of a single material throughout a natural scene.
Semantic and instance segmentation methods have a different goal: selecting entire objects.
In our material selection task, multiple objects can have the same material (e.g., multiple plastic chairs in a conference room), and a single object can be made of multiple materials (e.g., a chair with metal legs and leather seat).
Our method therefore adds a new dimension to image selections, allowing a user to easily select similar materials in an image.

\begin{figure*}
    \centering
    \includegraphics[width=\textwidth]{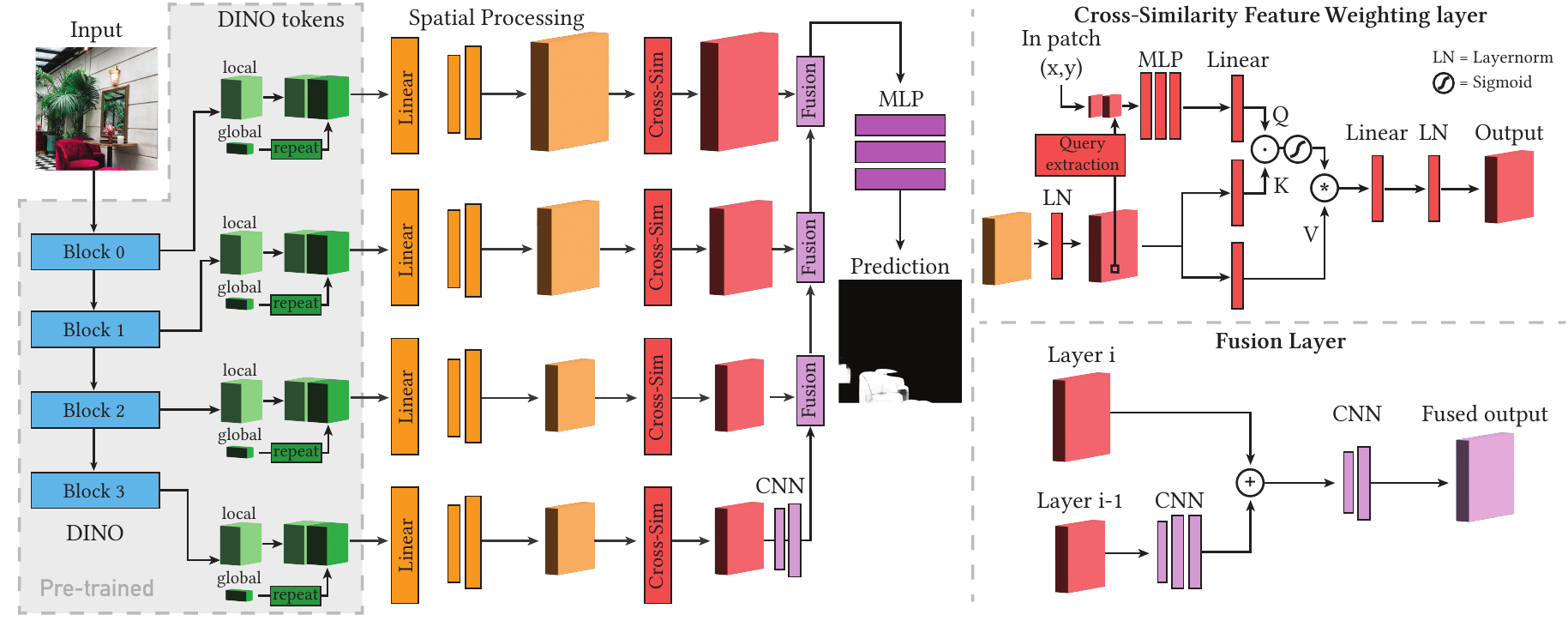}
    \caption{\textbf{Model Architecture.} We illustrate our model architecture in this figure. The DINO features extracted on the left come from a frozen, pre-trained DINO model. We then process these layers independently at different resolution, injecting the user reference through our \crosssim ~layer (Cross-Sim) before fusing the features and extraction the similarity prediction through a fully-connected head.}
    \label{fig:model_arch}
\end{figure*}

\section{Method}

% High-level goal
Given an input image and a query pixel location, our algorithm computes a scalar score that quantifies how similar materials at each pixel of the input image is to the material at the query pixel, from which we derive a binary material selection mask by thresholding.
The user can refine the selection by tweaking the threshold.

% Method roadmap
Our method, illustrated in Figure~\ref{fig:model_arch}, starts from rich self-supervised natural image features from a pretrained vision model (\S~\ref{sec:dino}), atop which we train a material-aware multi-scale encoder (\S~\ref{sec:spatial}).
This encoder serves two purposes.
First, it specializes the representation to be sensitive to material properties and insensitive to lighting, object-ness or other discriminative properties the pretrained features may contain. 
Second, it lets us refine the spatially coarse pretrained features into more precise per-pixel features.
We inject the spatial query point using a novel feature aggregation mechanism that weights the encoder's internal features
by cross-similarity at each scale of the encoder (\S~\ref{sec:crossim}).
We then fuse the multi-scale information before computing the final, per-pixel material similarity score (\S~\ref{sec:fusion}).
Because of the lack of high-quality, publicly available datasets with fine-grained material annotations,
we train our encoder on a new large dataset of photorealistic renderings of interior scenes with ground-truth material labels (\S~\ref{sec:dataset}).
Using features from a large pretrained vision model trained on natural images together with synthetic training data with ground-truth label gives us the best of both worlds: labels that are cheap to acquire, and straightforward generalization to natural images.
We provide the details of our network architectures in \ct{supplemental material}.

\subsection{Pre-trained DINO features}\label{sec:dino}
Self-supervised vision transformers (ViT) features have recently been shown to encode remarkable properties, such as rich semantic segmentation information~\cite{caron2021emerging}, which makes them a natural starting point for our material selection task.
As we discuss later (see Section~\ref{sec:dataset}), starting from features pretrained on natural images has the added advantage of mitigating the real--synthetic domain gap that often arises when training neural networks purely on synthetic data.

Concretely, we process our input image using DINO's ViT $8\times 8$ configuration~\cite{caron2021emerging} and extract a subset of its intermediate feature tensors.
Internally, DINO splits the image into non-overlapping $8\times 8$ patches, called `tokens', which are then processed by a series of transformer blocks.
Each block maintains a set of local tokens, which encode local patch information; and a global token that
encodes global context information~\cite{caron2021emerging,hamilton2022unsupervised}.
The DINO ViT model contains 12 attention blocks, we use the outputs of four blocks at index (2, 5, 8, 11) as our starting feature representation, inspired by ~\citet{DPT}.
We denote the local tokens, viewed as spatial feature tensors, by $\phi_i\in\mathbb{R}^{d\cdot\frac{h}{8}\cdot\frac{w}{8}}$, and the global tokens with $\psi_i\in\mathbb{R}^d$, where $h,w\in\mathbb{N}^2$ are the input's spatial dimensions, $d=768$ is the feature dimension, and $i\in\{1,\ldots,4\}$ indexes the blocks.
Because of the transformer's tokenization, the local feature tensors have $\frac{1}{8}$ the spatial resolution of the input image.

In section~\ref{sec:eval}, we show with an ablation that the DINO features significantly improve the selection quality, compared to a UNet-based baseline trained from scratch.

\subsection{Material feature encoder}\label{sec:spatial}
Despite their remarkable properties, the DINO features are generic and do not possess the invariants that make for a robust material representation.
Therefore, we transform them into material-specific features (\S~\ref{sec:multiscale}) by training a custom encoder on synthetic renderings with material ground-truth labels (\S~\ref{sec:dataset}).
We then condition the features on our query selection (\S~\ref{sec:crossim}), to finally compute our material similarity score (\S~\ref{sec:fusion}).

\subsubsection{Multi-scale features}\label{sec:multiscale}
Our encoder operates at multiple scales, to increase the spatial resolution of the DINO features and analyse multiple scales in the selection process.
We combine the global and local features and expand their spatial dimension following the pipeline of DPT~\cite{DPT}.
Specifically, for each transformer block $i=0,\ldots,3$, we first aggregate the local and global features by replicating the global token spatially and concatenating it with the local feature tensor.
Then, we process the aggregate using a convolutional network followed by a bilinear upsampling operator, with a different upscaling factor $s_i$ for each feature block, which yields a new set of features
\begin{equation}
     f_i(\phi_i,\psi_i) \in \mathbb{R}^{d'\times s_i\frac{h}{8}\times s_i\frac{w}{8}},
\end{equation}
with $d'=256$.
We use the following upsampling factors: $s_0 = 4$, $s_1 = 2$, $s_2 = s_3 = 1$, such that earlier feature blocks are upsampled more.

\subsubsection{Query injection using cross-similarity feature weighting}\label{sec:crossim}
After the convolutional stages described above, we obtained generic material features maps $f_i$ at resolutions $1/2, 1/4, 1/8, 1/8$ of the input image, respectively.
To implement a dynamic selection mechanism that does not rely on a predefined set of material classes, and can generalize to materials unseen during training, we need to transform $f_i$ into conditional features. These  conditional features must account for the material at the query pixel $q\in[0,1]^2$, using normalized coordinates, to simplify the multi-scale notation.

To do so, we propose a novel cross-similarity feature weighting operator that modulates the feature at another pixel $p\in[0,1]^2$ using a query-dependent weight.
We first obtain query Q, keys K and values V embeddings from $f_i$.
K and V are computed by processing $f_i$ with two different linear layers.
To obtain Q, we first extract the embedding at location $q$ from $f_i$ and concatenate the in-patch pixel coordinate of the query selection to it.
This provides our network with spatial information finer that the DINO patch index.
We then feed this embedding to an MLP that outputs Q.
The query-dependent weight of pixel $p$ at each resolution $i$ is then given by:
\begin{equation}
w_{i,pq} = \sigma(Q^T K / \sqrt{d}) \in [0, 1]^{s_i\frac{h}{8}\times s_i\frac{w}{8}}, 
\label{eq:sim}
\end{equation}
where $\sigma$ is a sigmoid activation.
Given this weight, we compute the weighted multiscale features to be fused as
\begin{equation}
g_{i,pq} = w_{i,pq}\cdot V.
\label{eq:weighted}
\end{equation}

Our feature weighting scheme is inspired by the attention mechanism~\cite{vaswani2017attention}, with a couple differences.
First, our similarity implements a one-to-many comparison, unlike the many-to-many relationship in traditional attention.
Second, we do not seek to compute relative importance in the feature map, but rather a non-negative similarity score between the query and all other embeddings.
So, the weights need not sum to one over the spatial dimensions, hence the use of a sigmoid in Eq.~\eqref{eq:sim} instead of the usual softmax.

\subsubsection{Multi-scale fusion and final material similarity score}\label{sec:fusion}

We progressively fuse the information from our query-conditioned multi-scale features $g_i$ from coarse to fine, using a residual network followed by $2\times$ bilinear upsampling between each consecutive scale, until we reach the full image resolution $h\times w$, 
Finally, we compute our material similarity score in $[0, 1]$ from the fused features using a pointwise neural network followed by a sigmoid activation.

\subsection{Datasets with fine-grained annotations}\label{sec:dataset}
As noted in Section~\ref{sec:RW}, existing material datasets~\cite{upchurch2022dense,bell2013opensurfaces,murmann2019dataset} with per-pixel material annotations contain \emph{semantic} material annotations.
These are too coarse for our application; they do not account for intra-class material variations.
For instance, two different wood types share the same label.
This prevents us from training and evaluating on these datasets for our task.
Instead, we rendered a synthetic dataset for training, and manually annotated a dataset of 50 real images for evaluation.

\begin{figure}
    \centering
    \includegraphics[width=0.3\linewidth]{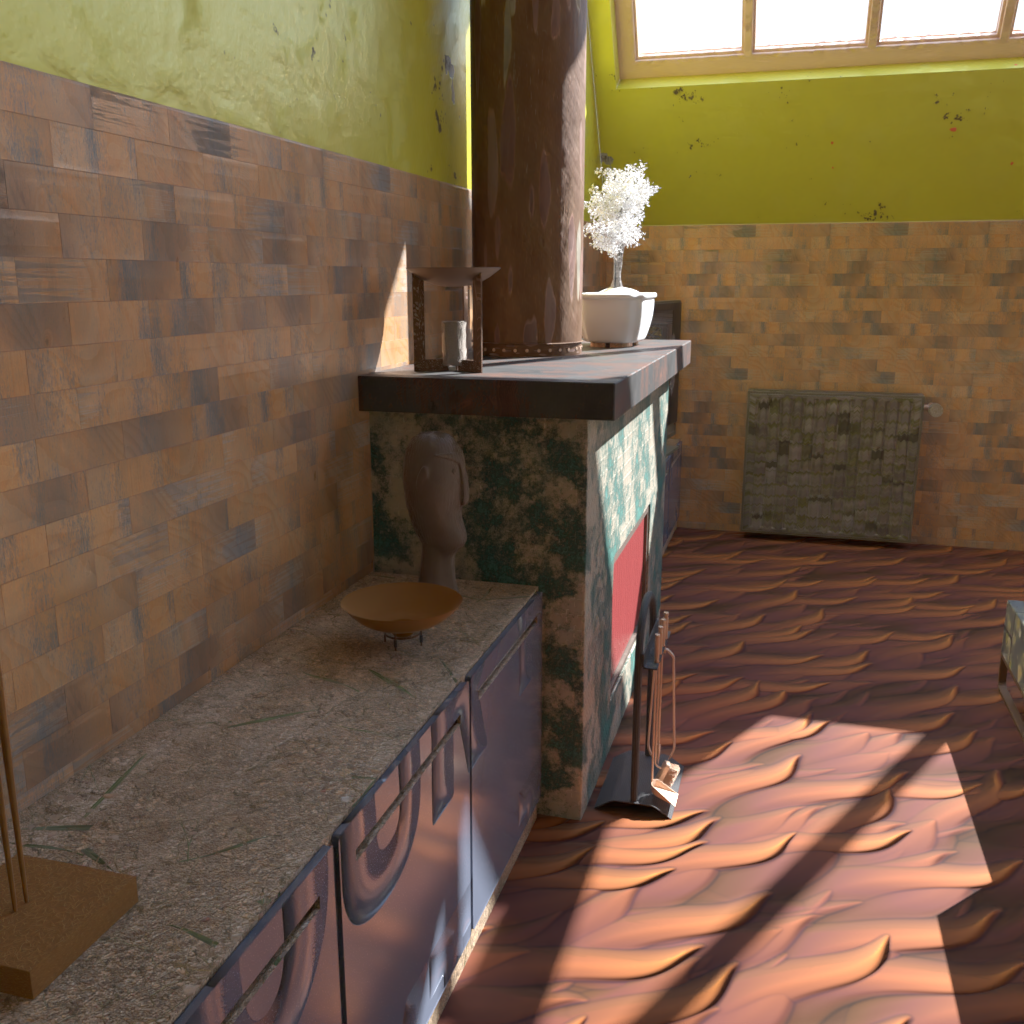}
    \includegraphics[width=0.3\linewidth]{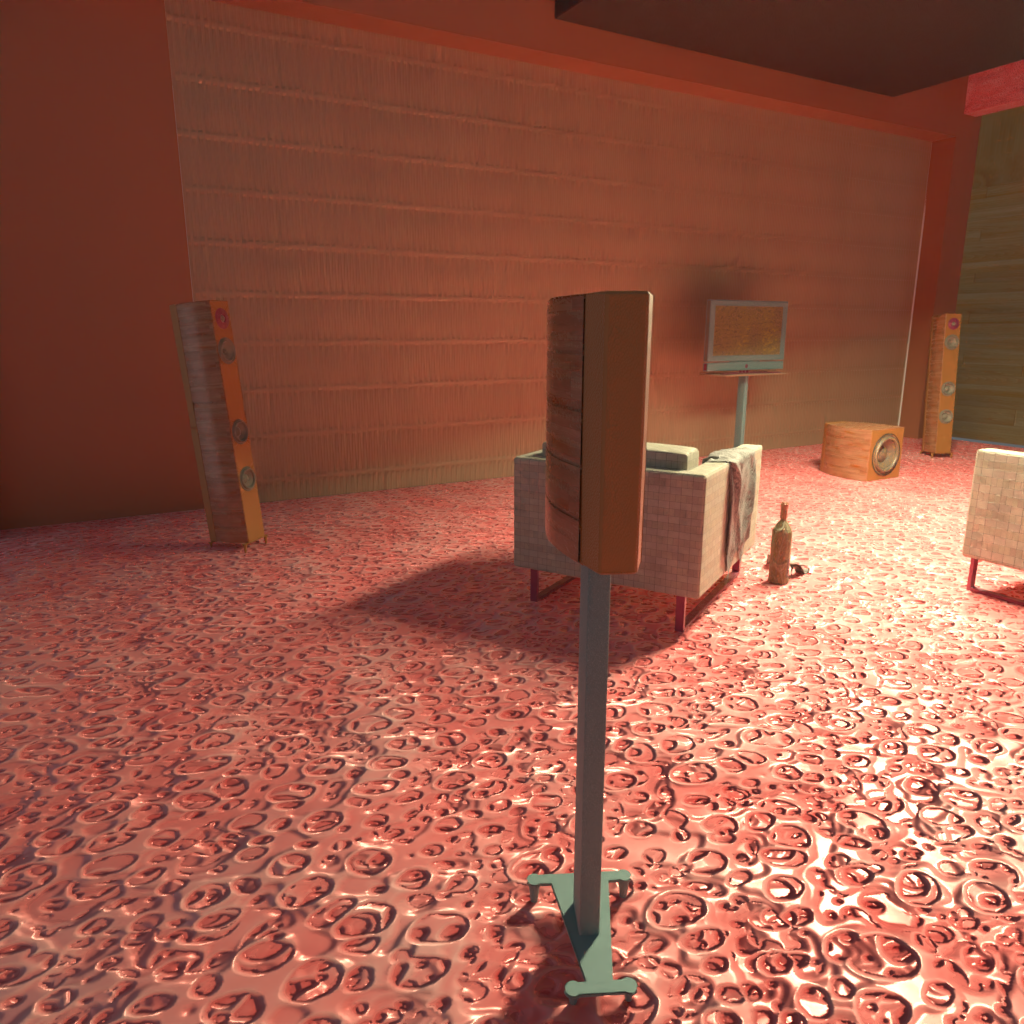}
    \includegraphics[width=0.3\linewidth]{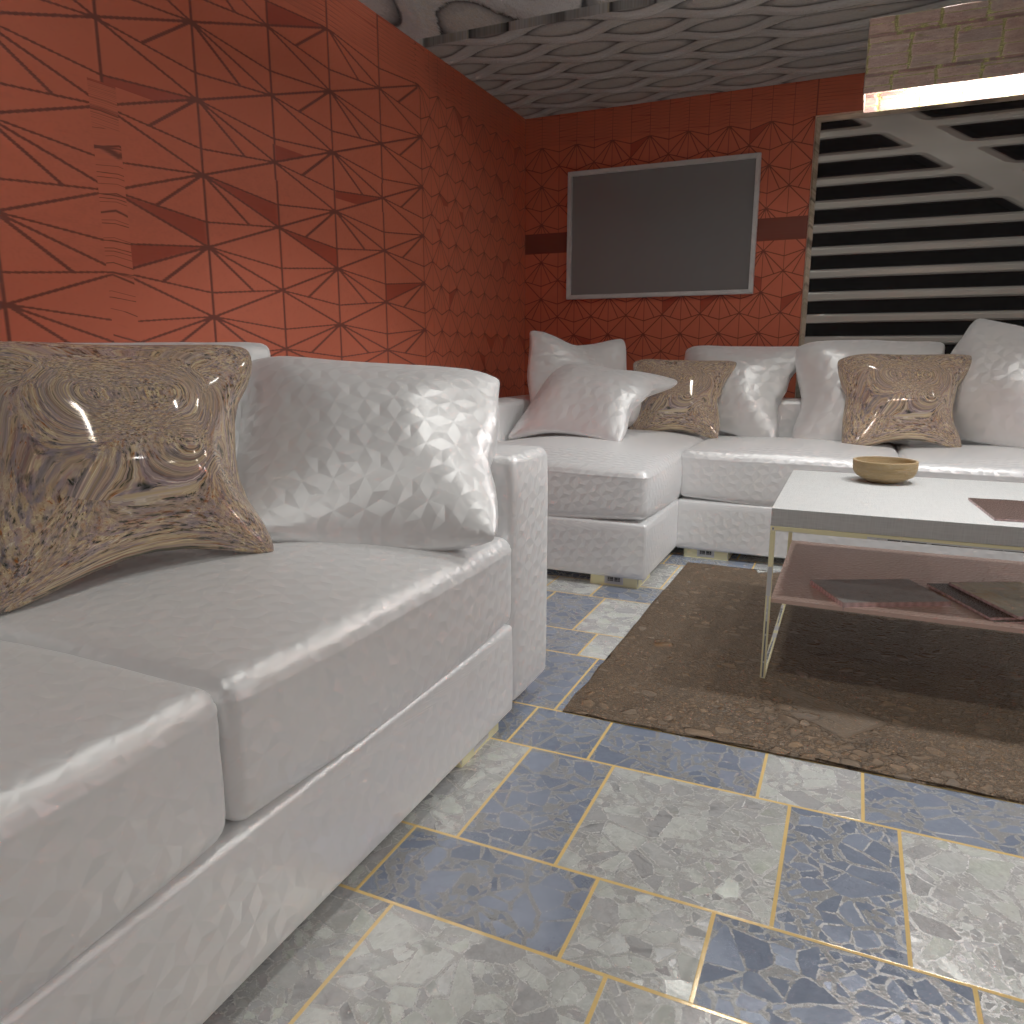}
    \includegraphics[width=0.3\linewidth]{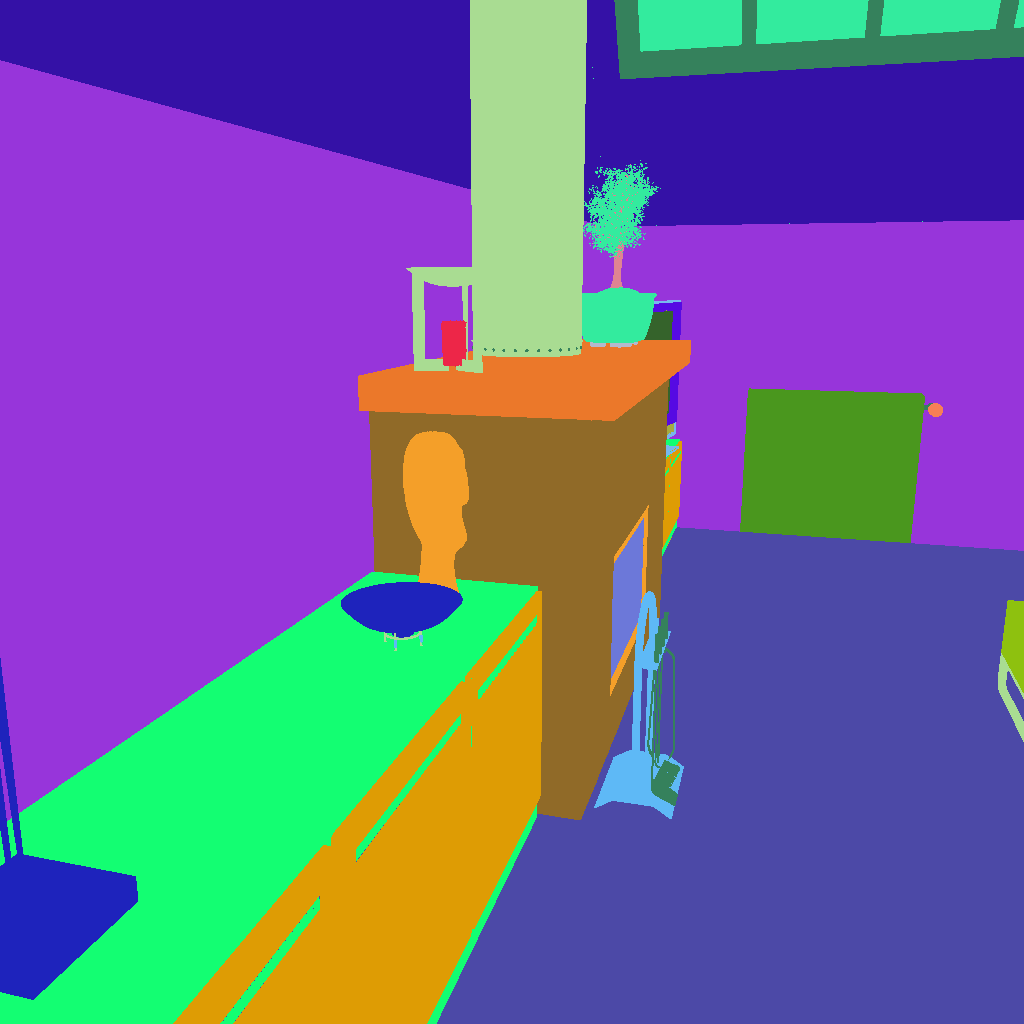}
    \includegraphics[width=0.3\linewidth]{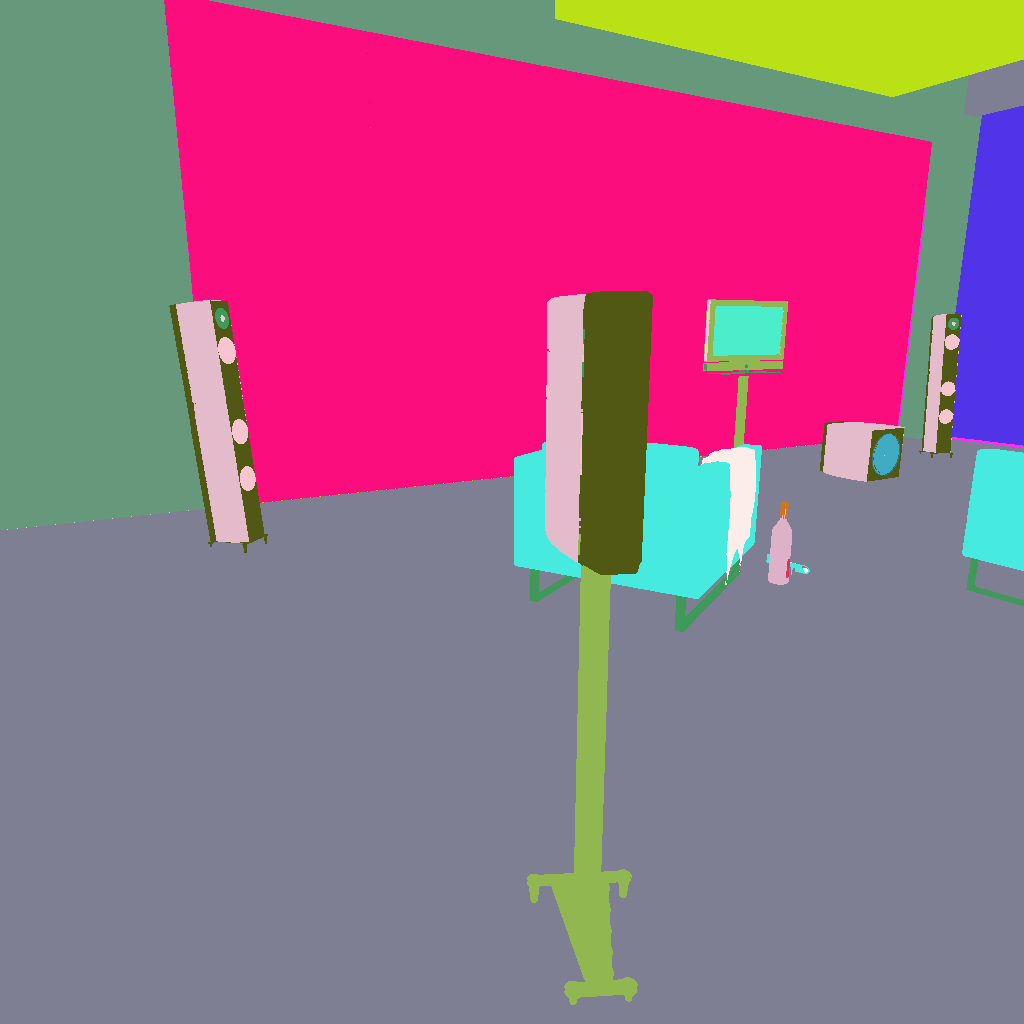}
    \includegraphics[width=0.3\linewidth]{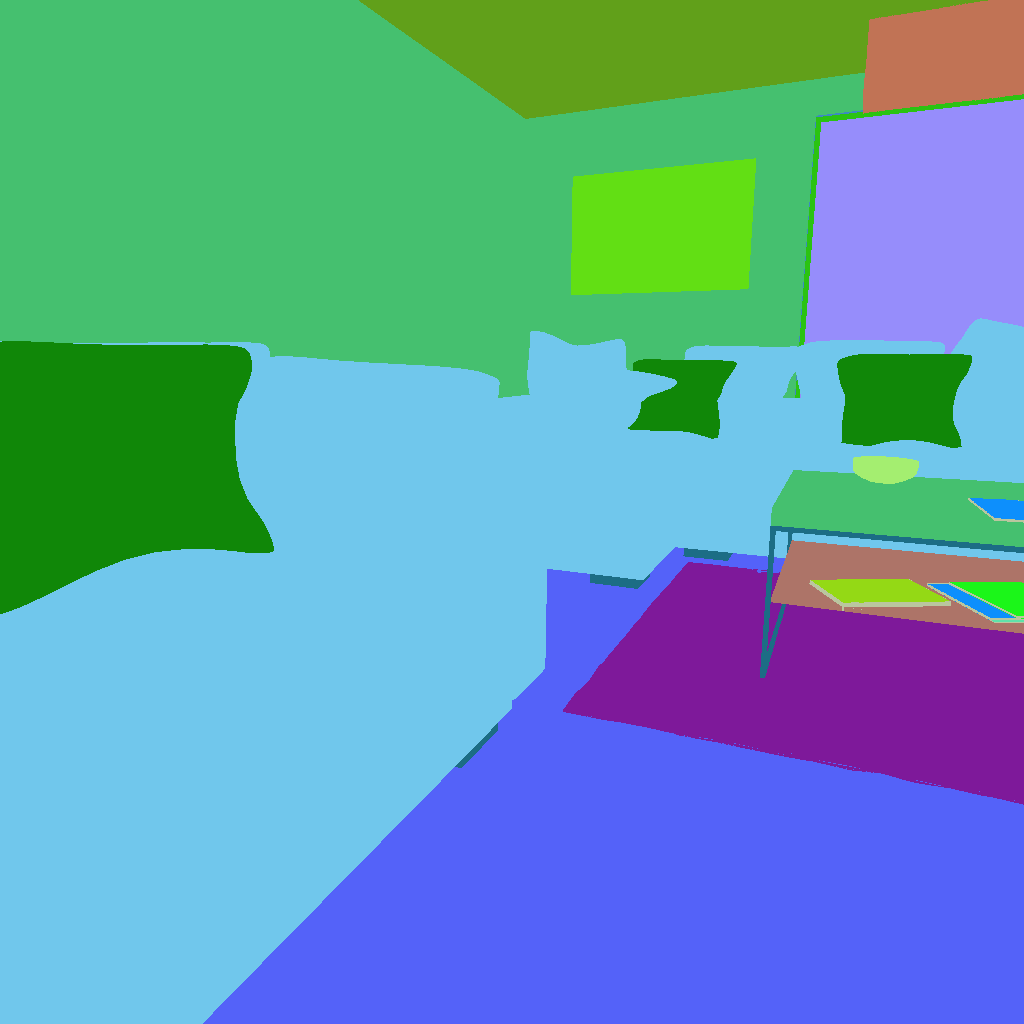}
    \caption{\textbf{Our synthetic dataset.} We show samples from our synthetic training dataset. Top: three sample rendered images. Bottom: The corresponding material id maps, ids are mapped to random colors.}
    \label{fig:dataset}
\end{figure}
\vspace{-4mm}

\subsubsection{Real-world evaluation benchmark}

For evaluation, we manually annotated 50 images using Label Studio~\cite{label_studio}, exhaustively segmenting on average 2 materials per image. The test images were sourced from Pixabay and Pexel and are available in the \ct{supplemental materials}.
They contain challenging cases such as multiple objects with the same material and objects made of multiple materials.
We will release this benchmark publicly.

\subsubsection{Synthetic training data}
Our synthetic training dataset is composed of $50,000$ HDR images rendered using Blender Cycles \shortcite{blender}.
For the geometry of our scenes, we use a subset of 100 scenes from the Archinteriors collection \cite{evermotion}. For each of the 100 scenes, we use a camera trajectory spline and an associated viewing direction spline that was recorded using a first-person exploration of the scene for about one minute. We randomly sample camera locations and associated directions from the path for each scene.
To render each image in the dataset, we randomly sample a camera position and field-of-view on a pre-defined, scene-specific camera path manually created using splines in Blender \shortcite{blender}.
For each rendering, we replace each material in the scene by randomly sampling a new one from a set of $3,000$ Adobe Substance Source materials which we curate to be stationary.
We keep the original material assignment of each object constant, meaning objects that share the same materials in the original scene still share the same material after replacement.

We show a few samples from our dataset in Fig.\ref{fig:dataset}.
The random combination of material and geometries dramatically increases the diversity of our dataset.
Replacing materials allows us to render an accompanying material ID map where the IDs are global across the dataset.
By using such ID maps for supervision, we implicitly define \emph{similar materials} as materials sharing the same stationary SVBDRF.
While we do not use the cross-image consistency of the IDs during training, we show in Figure~\ref{fig:cross_image} that our method is still capable making selections across multiple images. 
We render at a $1024\times 1024$ resolution with 256 samples per pixel using the original scene lighting.
Rendering the full dataset takes about 24 hours using 8 NVIDIA A10G GPUs.
Upon publication, we will release the $50,000$ HDR images and material ID pairs to facilitate further research on material selection.

\subsection{Implementation details}
We train our model for 30 epochs on the dataset described in Section~\ref{sec:dataset}, using the Adam optimizer with a learning rate of $10^{-4}$ on 2 V100 GPUs, with a batch size of 8 images per GPU, following the distributed data-parallel (DDP) training approach.
During training, we apply random exposure, saturation, and brightness augmentations to a random $512\times 512$ crop of our training renderings.
At inference time, our model computes material similarity in a $512\times 512$ image in 240 ms on a V100 GPU. 

\paragraph{Loss function.}
Given a query pixel, we compute a binary cross-entropy loss to measure whether the predicted selection (with threshold $0.5$) matches the desired material regions in the image, specified by the query pixel location and the ground truth material segmentation of our synthetic data. 

\paragraph{Selection refinement using KNN matting}
\label{sec:refinment_knn}
Because of the low resolution of the DINO feature, our approach may not always select the boundaries of thin features perfectly.
We found that applying the KNN-Matting algorithm~\cite{knn_matting} on the edges of our selection was sufficient to improve the resolution of small features, if desired.
When using this post-processing, we automatically define the positive and negative anchors required by KNN matting, by eroding and dilating our selection mask with a $(9, 9)$ kernel, marking the content of the erosion as positives, and the inverse of the dilation as negative anchors.
\emph{Unless explicitly specified, we show the direct output of our network and do not apply the refinement step discussed in here, which is separately illustrated in Figure~\ref{fig:comparison} and in the \ct{supplemental material}}.

\section{Results}
In this section, we present the qualitative results from our model, and its ability to perform material selection across images allowing us to directly apply our method to selection in high-resolution images and video frames for example. 

\begin{figure*}
    \centering
    \includegraphics[width=\linewidth]{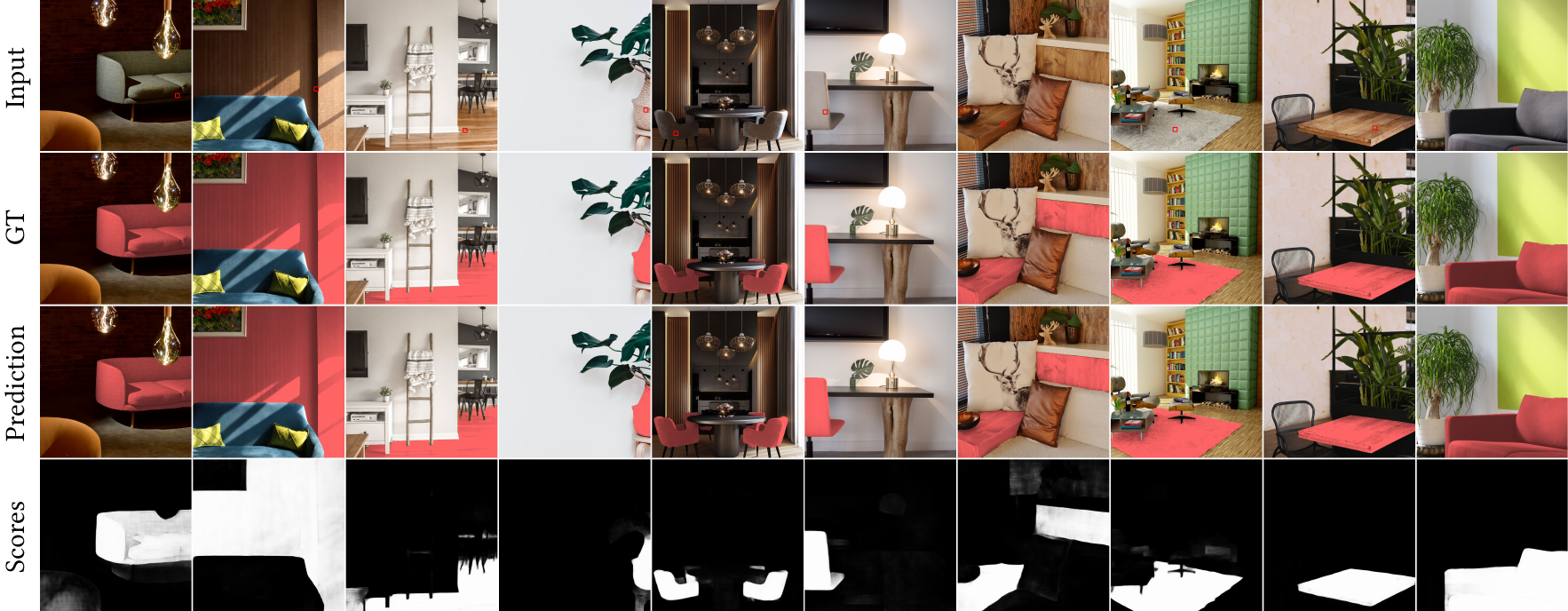}
     \caption{\textbf{Qualitative results.} We present the input image along with the query point annotated with a red square, ground truth mask (GT), predicted mask, and the per-pixel score (top to bottom rows). Note that the user selects the center of the marked square. The prediction is the best possible mask generated by optimally thresholding the per-pixel scores. The predicted results demonstrate the robustness to shading variations (first and second columns), the presence of different objects sharing the same material (fifth and seventh columns), and surface orientation (seventh column).}
    \label{fig:qual_eval}
\end{figure*}

\begin{figure}
    \centering
    \includegraphics[width=\linewidth]{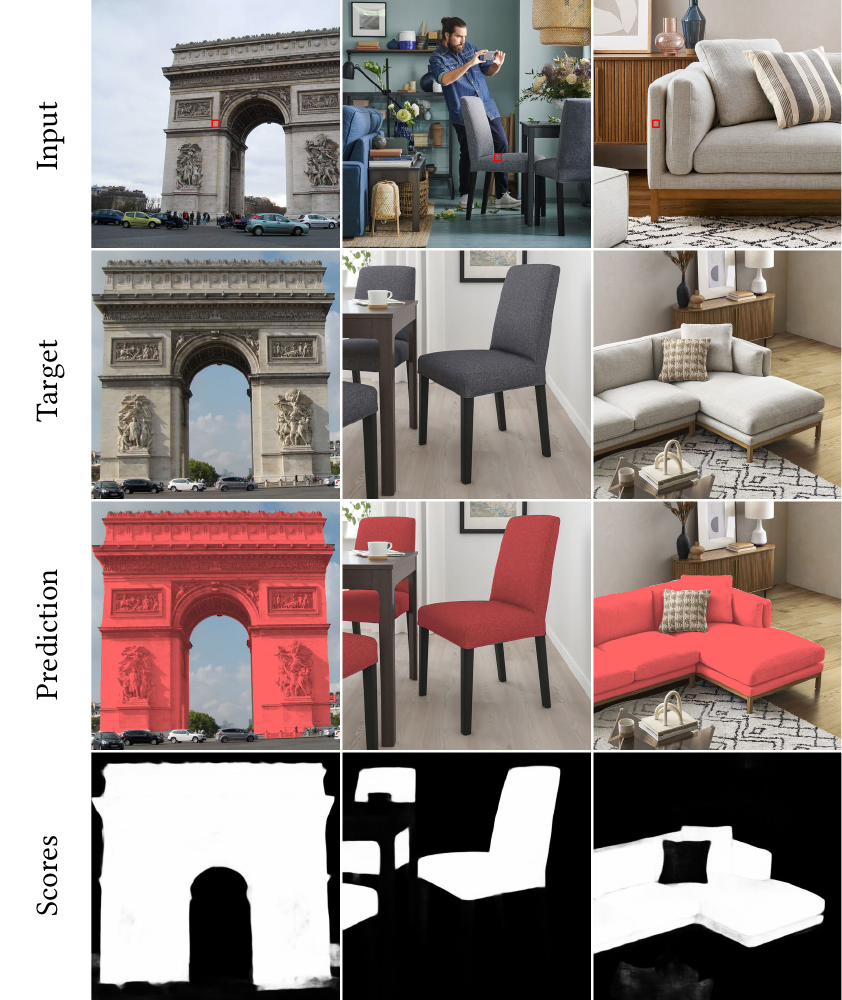}
    \caption{\textbf{Material selection across images.} From a selection in one image (top row), our model can find the corresponding material in a second image (second row). The reference pixels are highlighted with a red square and the predicted selection and associated similarity scores are shown in the third and fourth rows. We show that our model can indeed select across images, despite the differences in environment, viewpoint, and lighting.}
    \label{fig:cross_image}
\end{figure}

\begin{figure}
    \centering
    \includegraphics[width=\linewidth]{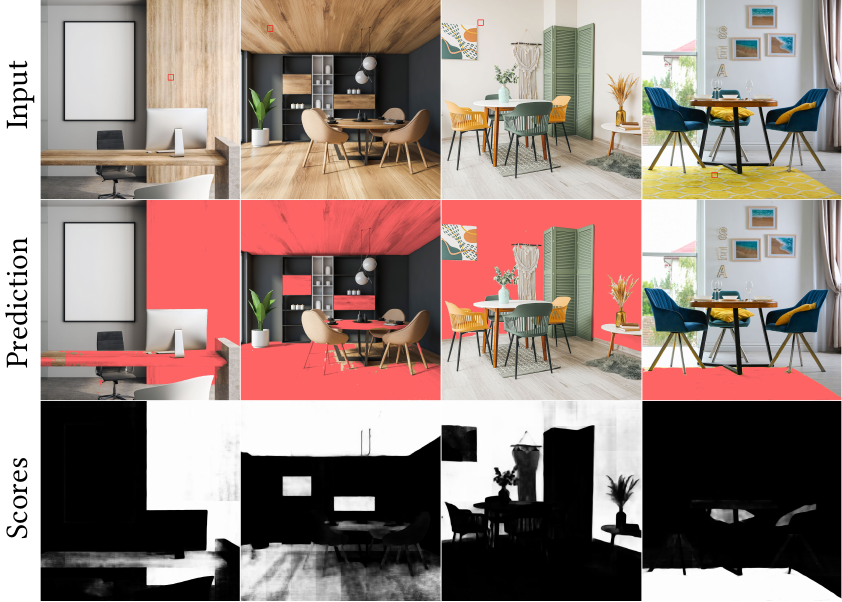}
    \caption{\textbf{High resolution material selection.} Our method can be evaluated on higher resolution such as $1024\times 1024$ even though the method was trained on $512\times 512$. Using the query embedding from a downsampled version of the image, we compute the similarity to the query of crops of the high-resolution image using a sliding window \update{with stride 256}. These scores coming from different patches are averaged for each pixel in the high-resolution image.}
    \label{fig:strided_1k}
    \vspace{-6mm}
\end{figure}

\begin{figure*}
    \centering
    \includegraphics[width=\linewidth]{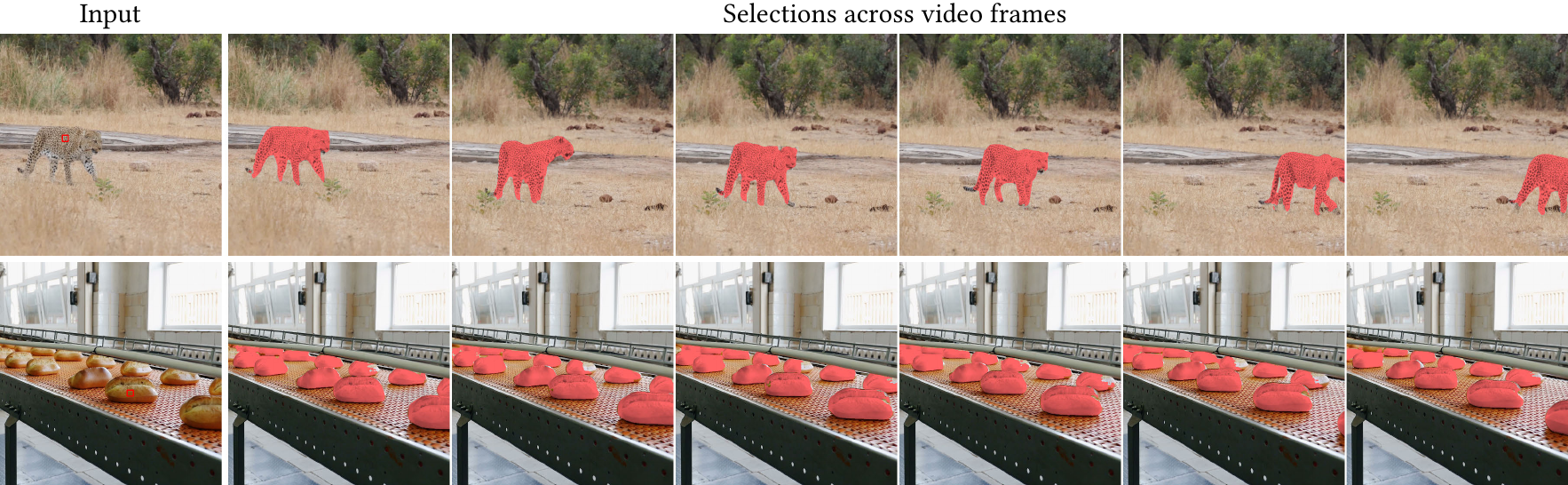}
    \caption{\textbf{Material selection in videos.} Given a user input on the first frame of the video, our method can select the material across all frames of the video. Note that the spotted fur of the cheetah is selected in all frames even though the dry grass texture also exhibits the same low frequency statistics. The method also successfully selects all the bread rolls on the assembly row.}
    \label{fig:video_results}
\end{figure*}

\subsection{Qualitative results} In Figure~\ref{fig:qual_eval}, we present results obtained by directly applying our method to a subset of our evaluation dataset.
This shows our model is robust to strong shading variations (first and second columns), to the presence of different objects sharing the same material (fifth and seventh columns), and to surface orientation (seventh column).
We also show in Figure~\ref{fig:comparison} (last column) that the KNN refinement step discussed in Section~\ref{sec:refinment_knn} can slightly improve selection boundaries and challenging thin structures.

\subsection{Cross-image selection}
\label{sec:across_images}
A natural extension of our method is the selection of similar materials across images. Given an image in which a user provides the query, we show that we can select similar materials in different images, as long as the lighting does not vary dramatically (we evaluate the robustness to lighting variation in Figure~\ref{fig:lighting_analysis}).

Given an input image $I_{query}$, used to define the user-query, we want to select similar materials in a different image $I_{select}$.
 To do so, we process both $I_{query}$ and $I_{select}$ independently up until the \crosssim{} layers. We then compute the query (Q in Figure~\ref{fig:model_arch}) using the features from $I_{query}$ while we use the spatially processed images features (K and V in Figure~\ref{fig:model_arch}) from $I_{select}$.
The cross-similarity features are then fused in the same way than for selection in a single image.

 We show cross-image selection results in Figure~\ref{fig:cross_image}. We can see that despite varying viewpoints and lighting conditions, our method can select similar materials across images. The ``Arc de Triomphe'' in the first column also illustrates that our method generalises well to outdoor images, further demonstrated in the video results discussed in Section~\ref{sec:video}, despite the training data being confined to indoor renderings.
 Note that even though the two images have different scene lighting, the model outputs robust selections.
 On the second and third column we observe that the materials from the chairs and sofa are well identified in the target image even though orientation and lighting vary.

 \subsection{Videos}
 \label{sec:video}
 The ability to select materials across images can also be used directly on videos where the reference pixel selection is made in the first frame, and the model selects material on a per-frame basis to extract the desired material across the video. We show results in Figure \ref{fig:teaser} and \ref{fig:video_results} and supplemental materials. These videos and Figure~\ref{fig:cross_image} further illustrate the stability and robustness of the approach even though no temporal smoothing is applied to the output.
 
\subsection{High resolution images}
The high computational cost of self-attention to compute DINO features restricts the input to our method to $512\times 512$ pixel resolution.
However, since our model generalizes across images, this lets us evaluate it on high-resolution images by processing overlapping crops.

Specifically, we follow the same process as cross-image selection.
We first obtain a query Q by running the model on a $512\times 512$ downsampled version of the image.
Then, using this query Q, we evaluate the similarity scores for a set of crops of size $512\times 512$ using a simple sliding window with a stride of 256 pixels on the high-resolution image. For each crop, we use their respective keys K and values V.

For each pixel in the high-resolution input we average the similarity scores obtained from all crops that overlap with this pixel, which gives us our final similarity score.
We present results on 1K resolution images in Figure~\ref{fig:strided_1k}.

\section{Evaluation}
\label{sec:eval}
\begin{table}
\caption{\textbf{Quantitative metrics.} Mean IoU scores of all models evaluated on our densely annotated material dataset containing 50 real images. \new{We compute the mIoU for 10 randomly selected user-query pixel for each image and average the results}. \ct{--see supplemental}. We also provide a tally which indicated if the method requires negative samples during inference.}
\begin{tabular}{@{}lcc@{}}
\toprule
\multicolumn{1}{c}{\textbf{Model}} & \multicolumn{1}{l}{\footnotesize{\textbf{Negative Samples}}} & \textbf{mIoU} $\uparrow$ \\ \midrule
KNN matting (3 patches)                       &                 \checkmark                              & 0.617         \\
KNN matting (5 patches)                       &                 \checkmark                              & 0.677           \\
KNN matting (3 patch, albedo estimates)                       &                 \checkmark                              & 0.567          \\
KNN matting (5 patch, albedo estimates)                       &                 \checkmark                              & 0.640         \\
\update{DMS~\cite{upchurch2022dense}}                       &                                              & \update{0.38} \\
UNet on RGB                        &                                               & 0.612        \\
DINO ViT16 backbone             &                            &   0.877       \\  
(Ablation) Single Dino Block       &                                               &    0.5      \\ 
(Ablation) No \update{Cross-Sim layer}       &                                               &    0.9      \\ 
(Ours) DINO ViT8 backbone         &                                               &    0.917      \\ 
Ours refined with KNNmatting               &                                               & \textbf{0.92}        \\ \bottomrule
\end{tabular}
\label{tab:miou}
\end{table}
\begin{figure*}
    \centering
    \includegraphics[width=\linewidth]{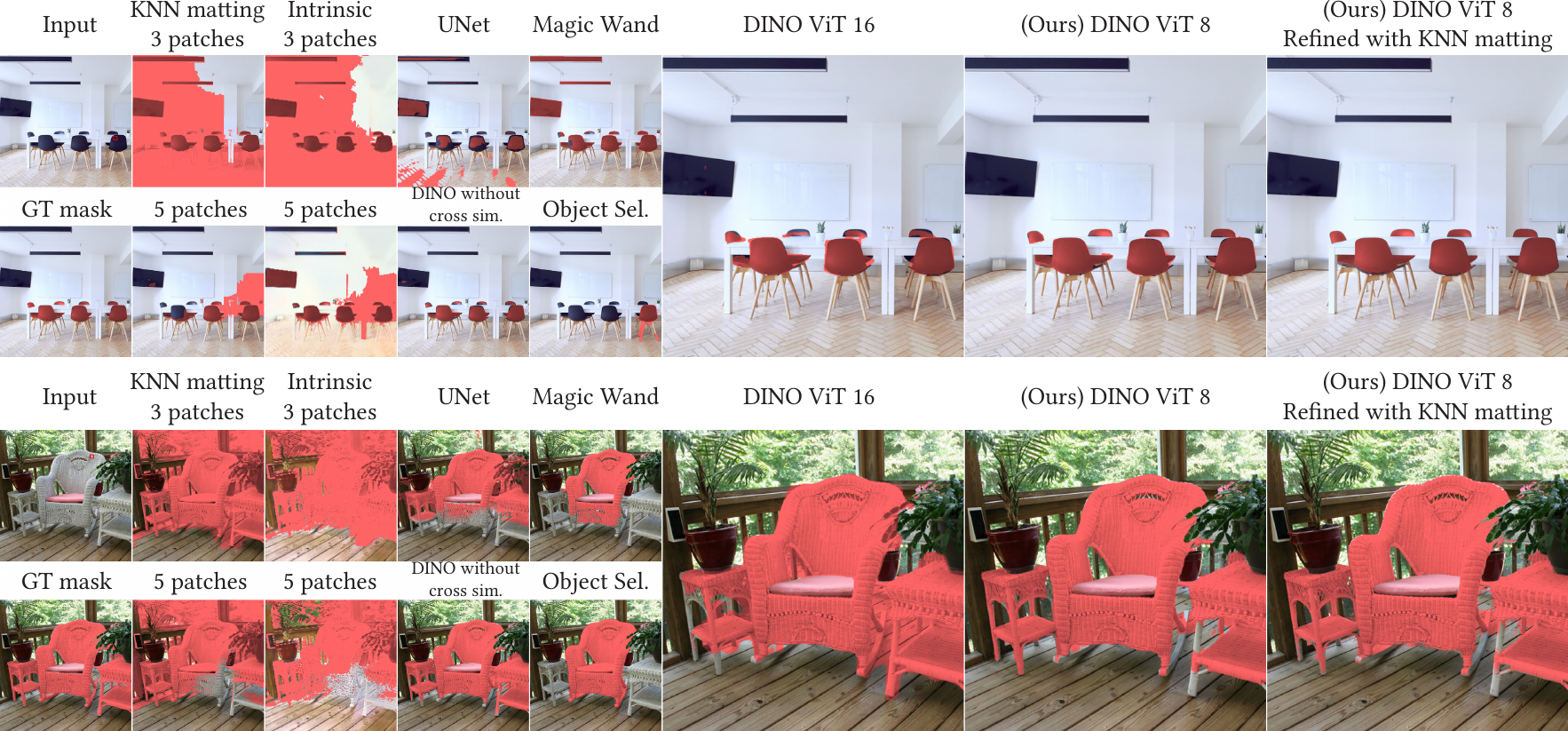}
    \caption{\textbf{Qualitative comparisons.} We show comparisons between selected regions (marked in red) obtained with different methods. From left to right, we first show results obtained with KNN matting using a different number of positive and negative patches, and in two different color domains (HSV and Intrinsic image albedo). We then show results for our two best-performing ablations, \emph{i.e.} the alternative UNet architecture and our architecture without the \crosssim{} mechanism. Then we present results from existing Photoshop selection tools and a VIT-16 pre-trained network.
    Finally, on the far right are our results and a KNN matting refined version of our results for which we refine selection borders. We can see that our method better selects  the same materials in the image across different objects, despite similar colors in the image (e.g color and lamps in the top row) or the space being cluttered (bottom row). Further, the KNN Matting improved selection better follows material edges. While close, the no-cross sim ablation has more imprecise selection and overselects the TV (top-row) and the background of the wooden chair (bottom-row).}
    \label{fig:comparison}
\end{figure*}

We present both quantitative and qualitative evaluations of our model through a set of ablations, comparisons and experiments to better understand its behavior and limitations. Quantitative results are shown in Table~\ref{tab:miou}, in which we provide the mean Intersection over Union (IoU) score comparing the ground truth material masks and the predicted masks on our benchmark evaluation dataset for 10 randomly selected query points per image.
We further show qualitative comparisons in Figure~\ref{fig:comparison}. As different methods require a different threshold, we always report the number with their optimal threshold selected. We perform a grid search to find the threshold that yields the highest mIOU for each method. This illustrates the best selection result that can be achieved using each method.
\subsection{Ablations}

\paragraph{Dense Material Segmentation}
We compare our method to a material segmentation method~\cite{upchurch2022dense}, where we run the pre-trained segmentation network on the test images and generating the binary mask as the segment that belongs to the segment at the user selection. Since this network is trained with a closed set of high-level material labels (i.e. wood, metal), the network suffers from under-segmentation as it treats all intra-class variations of the material as the same. 

\paragraph{UNet} 

We explore and evaluate an alternative neural architecture. Given the image to image nature of our task, we use the fixup UNet~\cite{ronneberger2015u, zhang2019fixup} which has been shown to do well on single-image relighting tasks~\cite{griffiths2022outcast}.
We train this model on the synthetic dataset presented in Section~\ref{sec:dataset}. Given an RGB input image $I$, the network outputs a per-pixel 32-dimensional embedding $f_p$. We train this approach with a binary cross entropy loss using the dot product between the query embedding $f_q$ and all other embeddings $f_p$.

This approach can be seen as an extremely simplified version of our model, where the query injection happens at a single resolution, and where the dot product between query and key (which are not processed through separate MLPs) is directly used as similarity instead of being further processed.
As shown in Table~\ref{tab:miou} this U-Net model achieves significantly lower mIoU, and we can see in Figure~\ref{fig:comparison} that the resulting selection tends to be noisy.

\begin{figure*}[h!]
    \centering
    \includegraphics[width=\linewidth]{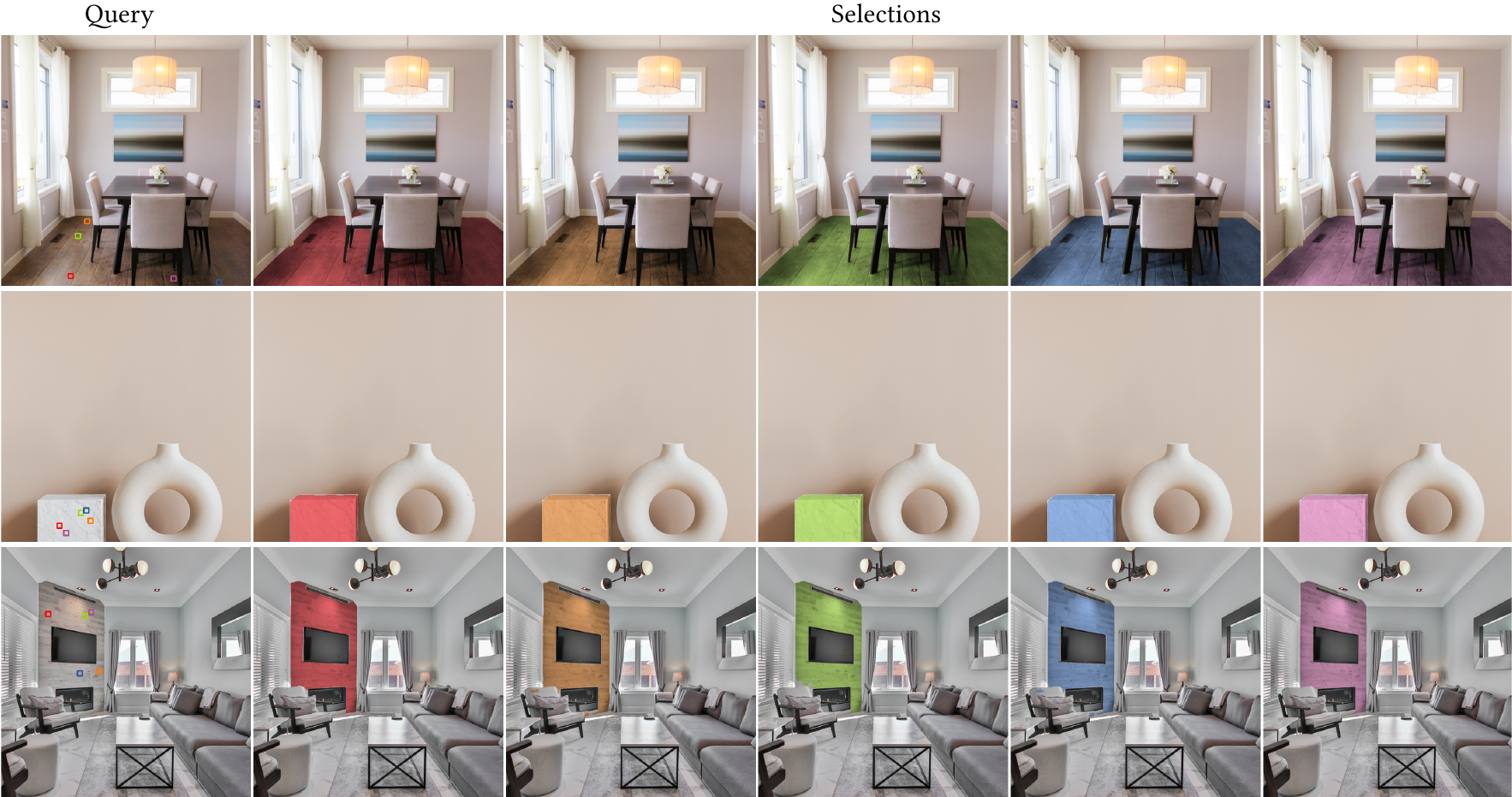}
    \caption{\textbf{Consistency of model output.} Querying our model with 5 different query points --highlighted by a colored square-- within region corresponding to the same material, the selected region is consistent. Each of the query points and the corresponding selection is presented in different colors. Note that despite the different lighting effects such as specularity and shadows around different selections, the model consistently selects the same region (row 1 and 3). Similar consistency is exhibited for selections around different regions of a spatially varying texture of marble in row 2.}
    \label{fig:selection_consistency}
\end{figure*}

\paragraph{Single DINO features block}
Our model uses multiple blocks of the pre-trained DINO model, which we need to fuse after injecting the query, as described in Sections~\ref{sec:crossim} and~\ref{sec:fusion}.
We evaluate the performance of a simplified model that uses a single DINO feature block.
For this baseline, we directly use $w_{3, pq}$ from Eq.\ref{eq:sim} as the similarity score, since no fusion is needed.
Table~\ref{tab:miou} shows a single feature block with limited processing is insufficient.
This confirms that the pre-trained DINO features are not natively sufficient to discriminate materials,
and that our model design, which refines the 4 blocks of DINO features with a query-dependent feature combination and selection mechanism, is essential.

\paragraph{DINO patch size}
To evaluate the impact of the patch sizes of the DINO features, we trained a variant of our model using the ViT-16 model as a backbone, using $16\times 16$ patches, instead of our main method, which uses ViT-8, i.e., $8\times 8$ patches.
The ViT-16 backbone performs better than the more naive baselines described above.
However, its lower spatial resolution leads to less precise segmentations, compared to the ViT-8.
This is especially visible around material edges in Figure \ref{fig:comparison}.
This also leads to a lower accuracy overall (see Table~\ref{tab:miou}).

\paragraph{Selection injection.}
Finally, we ablate our \crosssim{} layer, replacing it with
a simple concatenation, at every pixel, of the query features vector with the pixel's feature vector.
This new tensor is then processed by a linear layer before the fusion step. 
This simplified network, denoted ``(Ablation) No Cross-Sim layer'' in Table~\ref{tab:miou}, performs reasonably well.
But removing our feature reweighting and the injection of the local sub-patch position of the query pixel (Figure~\ref{fig:model_arch} top-right), leads to degraded spatial localization.
This is especially visible in Figure~\ref{fig:comparison}, where the selection is imprecise around the chairs edges and some part of the TV (top-row).
The baseline also overselects the wooden chair background (bottom row).
More comparisons are provided in the \ct{supplemental material}.
\vspace{-4mm}
\subsection{Comparisons}
In addition to these ablations, we compare to existing selection methods.
Our method being the first to enable material selection for natural images, we compare our model to the following selection tools: KNN matting~\cite{knn_matting} in multiple color spaces (HSV, albedo from intrinsic images\cite{li2020inverse}) and the Magic Wand and Object selection tools in Photoshop.
For all comparisons and ablations, we show additional results in \ct{supplemental material}

\paragraph{KNN matting} We use the open-source implementation of KNN matting~\cite{pymatting} on our test data in HSV space~\cite{knn_matting}.
The method requires positive and negative anchors, so we report their results using 3 and 5 $32 \times 32$ patches as positive and negative samples.
We select the patch randomly within the ground-truth positive/negative regions.
We also evaluate KNN matting on an albedo map extracted using intrinsic image decomposition~\cite{li2020inverse}, to try and minimize the effect of shading in the selection.

Based on the mean IoU scores in Table~\ref{tab:miou}, our method outperforms the baseline models. 
Since KNN matting only takes into account the observed color and the local pixel position, without any notion of lighting and geometry, its selection degrades when cast shadows and light dependent effects are present.
While the albedo component extracted from an intrinsic image decomposition method should remove shading , current methods do not handle global illumination effects perfectly~\cite{garces2022survey}, leading to artefacts in the image.
Moreover, recent intrinsic image decomposition methods output low resolution albedo maps.
Together, this limits the quality of KNN matting on intrinsic images baseline, so that it is no better than
running the algorithm in its original HSV space.

\paragraph{Photoshop selections}
We also compare our method qualitatively to results obtained using existing selection tools, namely the Magic Wand and Object selection tools in Photoshop.
Object selection works well but it solves a different task than ours, typically selecting a single entire chair, and not just the seat, in both examples of Figure.~\ref{fig:comparison}. 

On the other hand, Magic Wand selection is based on color similarity and often selects incorrect areas due to shading or specular highlights. As we show in Figure~\ref{fig:comparison} it does not handle well shading variations and varying lighting.
Further, similar RGB color (such as the TV and lights in the first row) are also incorrectly selected.

\subsection{Selection consistency}
We study the consistency of our model's output with respect to change in the scene lighting and selected pixel location.

\paragraph{Robustness to the query pixel.}
\label{par:query_pixel}
We evaluate the self-consistency of our model's selections by making multiple queries within a region labeled as a single material.

Figure~\ref{fig:selection_consistency} illustrates this experiment for 5 different selections marked in the input with squares of different colors along with the corresponding output selections.
Regardless location of the query pixel within a given material's region, our method's output selection
is stable.
In particular, as shown in rows 1 and 3, our selection does not change, even if the query points are under different illuminations.
We compute cross-mIoU between the selected regions predicted with different input query points within the regions corresponding to the same material. We consider the first selection point to be the control and compute the average mIoU of 5 other selections with respect to control. We compute this over our entire benchmark dataset with each image evaluated twice with random selection of query points and obtain an average cross-mIoU of 0.9387.

\begin{figure}
    \centering
    \includegraphics[width=\linewidth]{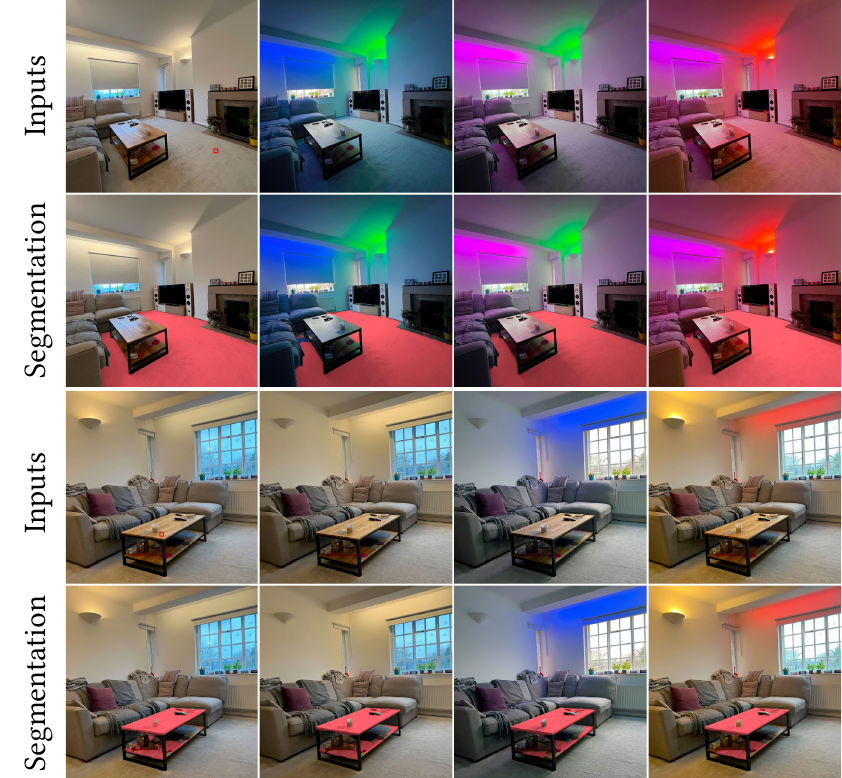}
    \caption{\textbf{Lighting Robustness Analysis.} We show the consistency of the selection of a given material for scenes with varying lighting.
    }
    \label{fig:lighting_analysis}
\end{figure}

\paragraph{Consistency under varying lighting.}

Previous sections showed qualitatively that our method is robust to varying lighting.
Here, we evaluate this systematically.
In Figure~\ref{fig:lighting_analysis}, we show a room photographed multiple times, from the same viewpoint, each time changing the color and intensity of the room's artificial lighting.
For this test, we used the ``across images'' approach described in Section~\ref{sec:across_images}.
The query is computed from the leftmost image in each row and then used to select the same material in the other images with the same viewpoint.
Our selections are fairly robust to strong lighting variations, although slight variations can be observed in areas where the radiance changes significantly, such as the grey carpet turning deep blue in the first example.

We show further examples with varying lighting from the dataset of \citet{murmann2019dataset} in \ct{supplemental material}.
We compute the cross-mIoU again (as described in \S\ref{par:query_pixel}) between the selections across different lighting conditions for the photographs shown in Figure~\ref{fig:lighting_analysis} and examples from \citet{murmann2019dataset} dataset shown in the supplement material. Here we obtain an average cross-mIoU of $0.956$. 

\subsection{Varying color and texture}

Color is a very strong visual cue, on which many segmentation methods rely (e.g., Photoshop's Magic Wand or KNN Matting~\cite{knn_matting})).
To evaluate the importance of color and texture in our model's selection, we designed two synthetic test cases, in which color (resp.\ texture) varies progressively across multiple spheres (see \ct{supplemental material.}).
In particular, this test helps understand what the method considers to be a single material.
Our model behaves as expected, with no color variations tolerated in the selection at high threshold. 
As we lower the threshold, the selection is relaxed and some color variation is tolerate.
Similarly, in the texture interpolation test, we see that closer interpolated textures are selected first, with a loose threshold.

\subsection{Refining selections with multiple query points}

To further empower artists, in our interactive demo, we allow users to select multiple positive and negative query points.
The resulting score map for positive query points are combined by taking the maximum of the individually predicted similarity scores for each pixel, and thresholded with a user defined value in $[0, 1]$.
Similarly, the user can select negative points to remove regions from the current selection. The predicted scores corresponding to all negative samples are also combined by computing a per-pixel maximum across all predicted scores, and then thresholded by the user using a separate threshold value.
The intersection of the negative mask with the mask computed using positive query points is removed from the final selection. We illustrate this workflow in a video in \ct{supplemental material}.

\begin{figure}
    \centering
    \includegraphics[width=\linewidth]{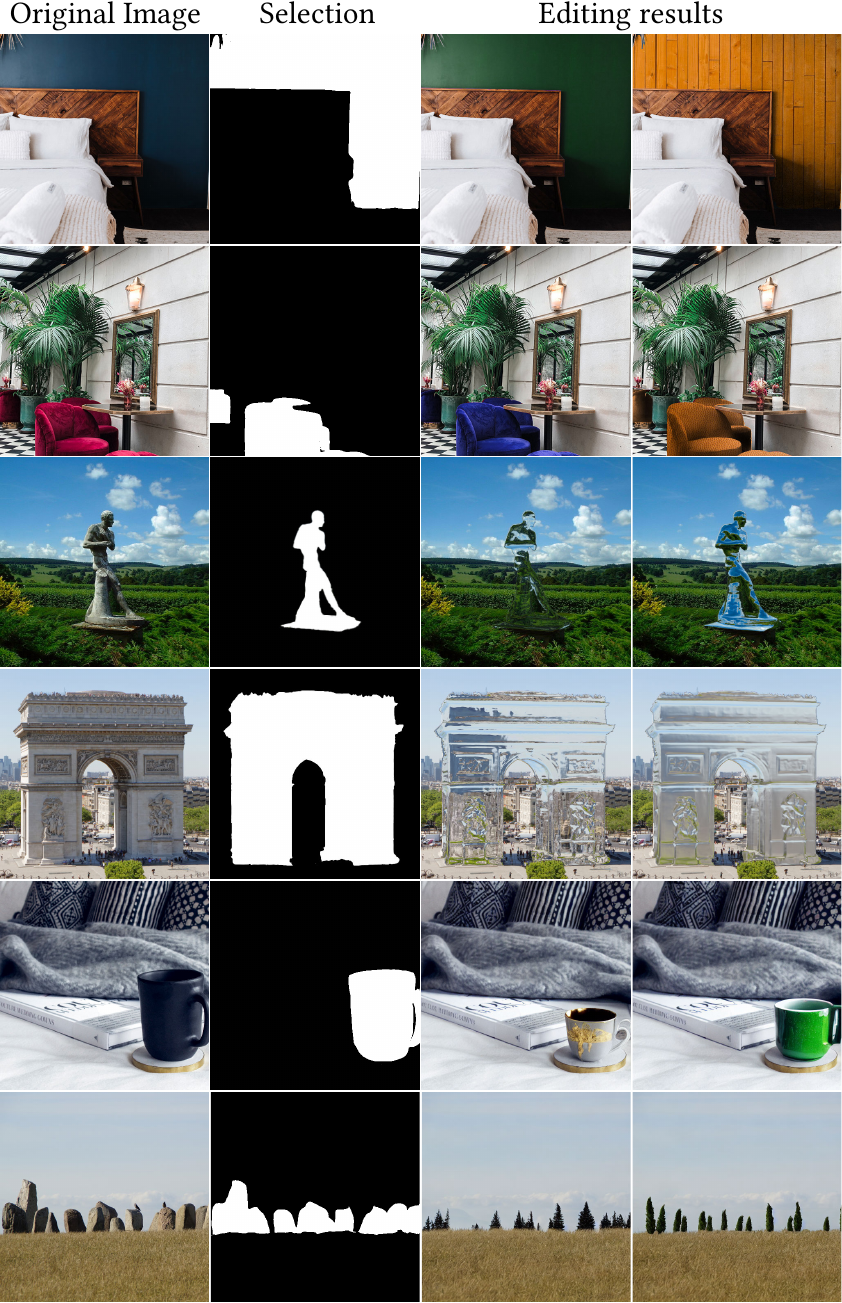}
    \caption{\textbf{Image editing results.} Our selection method output used as input for material selection based image editing using e.g. Photoshop (top rows), Khan et al.~\shortcite{10.1145/1141911.1141937} (middle rows) and Stable Diffusion Inpainting~\cite{rombach2021highresolution} (bottom rows). \new{Fifth row prompts: "A white tea cup with gilding", "A green mug". Last row prompts: "Pine trees", "Mediterranean trees".}}
    \label{fig:imedit}
\end{figure}

\section{Applications}

We showcase two applications of our model: image editing and material-driven web recommendations.

\paragraph{Image editing.} The ability to make selections based on materials opens up many image editing possibilities.
For instance, Figure \ref{fig:imedit} shows examples in which we alter or replace the selected materials.
In the top two rows, we modify the hue of the selection (first result column), and multiply the luminance with texture to replace the selection's appearance (second result).
For the next two rows, we partially re-implement the material-editing method from \citet{10.1145/1141911.1141937} using contemporary techniques, such as monocular depth estimation~\cite{Ranftl2022} and GAN-based inpainting for the environment~\cite{karimi2022ImmerseGAN}.
The first edit for the statue (3rd row) uses their glass approximation, while the second uses a pure mirror material.
In the Arc de Triomphe example (4th row), we use the glass approximation with varying roughness, implemented by blurring the environment map.
The last two rows are obtained using our selection as inpainting mask in Stable Diffusion~\cite{rombach2021highresolution} and various text prompts. 

\paragraph{Recommendation based on a selected material.}
Large online datasets such as an online product catalog can be challenging to navigate. Using our method we show that we can add a new axis along which it is possible to explore the dataset: material similarity. Given a subset of 150 images for different semantic material classes (wood, plastic, leather), from the Amazon Berkeley Objects (ABO) Dataset~\cite{collins2022abo}, we first select an image with a material we want to see more of, and search for objects with similar material using our query embedding following the material selection across images approach (Section \ref{sec:across_images}). We rank the images in the database's subset based on the mIoU of the selection with
respect to object mask in the rest of the dataset. As shown in Figure~\ref{fig:web_rec}, our method can retrieve objects with similar materials, we show the complete subset in \ct{supplemental material}.

\begin{figure}
    \centering
    \includegraphics[width=\linewidth]{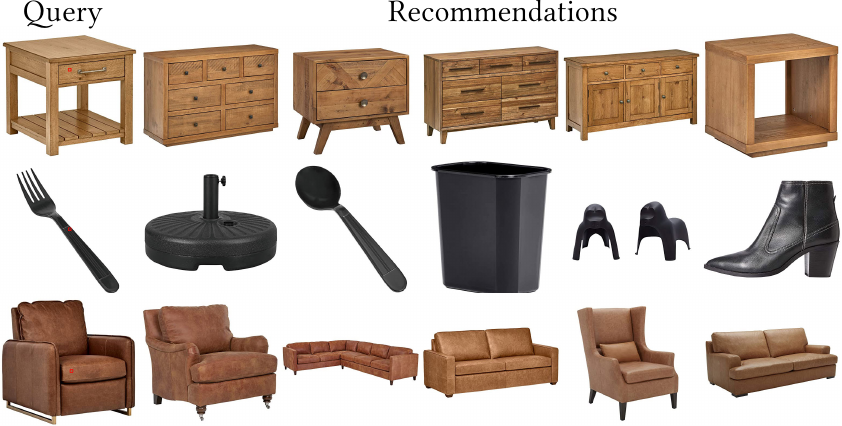}
    \caption{\textbf{Web recommendation.} Our method can be employed to perform material based search for web recommendation. Given a query image and a user selection, we rank the images based on mIoU of the selection with respect to object mask in a random 150 images subset of the Amazon Berkeley Dataset~\cite{collins2022abo} from different (wood, plastic leather). We show the complete subset in the \ct{supplemental material}.}
    \label{fig:web_rec}
\end{figure}

\subsection{Limitations}
\begin{figure}
    \centering
    \includegraphics[width=\linewidth]{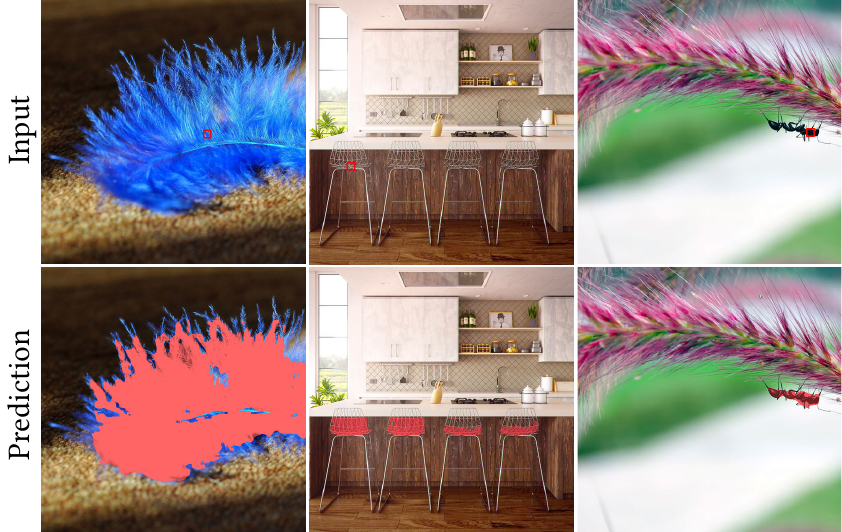}
    \caption{\textbf{Limitations.} The first two images illustrate the challenge of selecting thin structures for our method, while the last image illustrates the difficulty of extracting precisely high frequency and small selection borders.}
    \label{fig:limitations}
\end{figure}

As shown in the evaluations, our method is robust to light and view variations.
Despite being trained on purely synthetic data, it generalizes to real photographs and unseen materials, which in turns enables diverse applications. 

However, as mentioned in Section~\ref{sec:refinment_knn} selections of fine details remain challenging.
As shown in Figure~\ref{fig:limitations}, thin elements such as the blue feather, the thin grid on the chair or the ant are very hard to segment accurately.
Our KNN refinement step helps clean the selection boundaries in difficult cases, but is insufficient to recover very thin structures.
We believe stems from two limitations: the low resolution DINO features, mitigated but not entirely solved by our rescaling and feature-weighting mechanisms; and our synthetic training data, which does not contain many thin geometric structures. 

At a higher level, our definition of what constitutes a single material is closely tied to the notion of material in Computer Graphics, and what artists commonly define as stationary materials.
For example, we consider a wood plank as a single material, despite the wood-growth hue variations.
This definition may not always align with a user's expectation, but a different definition may require more fine-grained ground truth labels.
Our method is inaccurate for regions with extreme direct cast shadows, as seen in the second example presented in Fig 6. The direct cast shadows in such cases result in extremely underexposed regions revealing very little about the material in that region.

\section{Conclusion}
In summary, we propose a method for material selection in natural images.
Our method builds on pre-trained generic vision features, which we specialize for material selection by training a downstream model on a new synthetic dataset.
Crucially, our downstream model employs a new mechanism to merge multi-scale features and inject a user input.
We demonstrate the quality and robustness of our selections on both indoor and outdoor scenes, and show it can be applied to make selections with single images, across multiple images or even in videos.
We believe our method enables better high-level scene understanding and provides important information for inverse rendering optimization.

\begin{acks}
This work is in part supported by NSF 1955864, Occlusion and Directional Resolution in Computational Imaging, NSF under Cooperative Agreement PHY-2019786 (The NSF AI Institute for Artificial Intelligence and Fundamental Interactions, \href{http://iaifi.org/}{http://iaifi.org/}) and The Toyota Research Institute. This article solely reflects the opinions and conclusions of its authors and not other entity.
\end{acks}

\bibliographystyle{ACM-Reference-Format}
\bibliography{main}


\begin{thebibliography}{92}

%%% the Web address.

\ifx \showCODEN    \undefined \def \showCODEN     #1{\unskip}     \fi
\ifx \showDOI      \undefined \def \showDOI       #1{#1}\fi
\ifx \showISBNx    \undefined \def \showISBNx     #1{\unskip}     \fi
\ifx \showISBNxiii \undefined \def \showISBNxiii  #1{\unskip}     \fi
\ifx \showISSN     \undefined \def \showISSN      #1{\unskip}     \fi
\ifx \showLCCN     \undefined \def \showLCCN      #1{\unskip}     \fi
\ifx \shownote     \undefined \def \shownote      #1{#1}          \fi
\ifx \showarticletitle \undefined \def \showarticletitle #1{#1}   \fi
\ifx \showURL      \undefined \def \showURL       {\relax}        \fi
% invisible to TeX
\providecommand\bibfield[2]{#2}
\providecommand\bibinfo[2]{#2}
\providecommand\natexlab[1]{#1}
\providecommand\showeprint[2][]{arXiv:#2}

\bibitem[\protect\citeauthoryear{??}{eve}{2021}]%
        {evermotion}
 \bibinfo{year}{2021}\natexlab{}.
\newblock \bibinfo{title}{Evermotion Arch Interior}.
\newblock
\newblock
\newblock
\shownote{\url{https://evermotion.org/shop/cat/397/archinteriors}.}


\bibitem[\protect\citeauthoryear{Adelson}{Adelson}{2001}]%
        {adelson2001seeing}
\bibfield{author}{\bibinfo{person}{Edward~H Adelson}.}
  \bibinfo{year}{2001}\natexlab{}.
\newblock \showarticletitle{On seeing stuff: the perception of materials by
  humans and machines}. In \bibinfo{booktitle}{\emph{Human vision and
  electronic imaging VI}}, Vol.~\bibinfo{volume}{4299}. SPIE,
  \bibinfo{pages}{1--12}.
\newblock


\bibitem[\protect\citeauthoryear{Aittala, Aila, and Lehtinen}{Aittala
  et~al\mbox{.}}{2016}]%
        {Aittala2016}
\bibfield{author}{\bibinfo{person}{Miika Aittala}, \bibinfo{person}{Timo Aila},
  {and} \bibinfo{person}{Jaakko Lehtinen}.} \bibinfo{year}{2016}\natexlab{}.
\newblock \showarticletitle{Reflectance Modeling by Neural Texture Synthesis}.
\newblock \bibinfo{journal}{\emph{ACM Trans. Graph.}} \bibinfo{volume}{35},
  \bibinfo{number}{4} (\bibinfo{year}{2016}), \bibinfo{pages}{65:1--65:13}.
\newblock


\bibitem[\protect\citeauthoryear{Aittala, Weyrich, and Lehtinen}{Aittala
  et~al\mbox{.}}{2015}]%
        {Aittala2015}
\bibfield{author}{\bibinfo{person}{Miika Aittala}, \bibinfo{person}{Tim
  Weyrich}, {and} \bibinfo{person}{Jaakko Lehtinen}.}
  \bibinfo{year}{2015}\natexlab{}.
\newblock \showarticletitle{Two-shot SVBRDF Capture for Stationary Materials}.
\newblock \bibinfo{journal}{\emph{ACM Trans. Graph.}} \bibinfo{volume}{34},
  \bibinfo{number}{4} (\bibinfo{year}{2015}), \bibinfo{pages}{110:1--110:13}.
\newblock


\bibitem[\protect\citeauthoryear{An and Pellacini}{An and Pellacini}{2008}]%
        {An08}
\bibfield{author}{\bibinfo{person}{Xiaobo An} {and} \bibinfo{person}{Fabio
  Pellacini}.} \bibinfo{year}{2008}\natexlab{}.
\newblock \showarticletitle{AppProp: All-Pairs Appearance-Space Edit
  Propagation}. In \bibinfo{booktitle}{\emph{ACM SIGGRAPH 2008 Papers}} (Los
  Angeles, California) \emph{(\bibinfo{series}{SIGGRAPH '08})}.
  \bibinfo{publisher}{Association for Computing Machinery},
  \bibinfo{address}{New York, NY, USA}, Article \bibinfo{articleno}{40},
  \bibinfo{numpages}{9}~pages.
\newblock
\showISBNx{9781450301121}
\urldef\tempurl%
\url{https://doi.org/10.1145/1399504.1360639}
\showDOI{\tempurl}


\bibitem[\protect\citeauthoryear{Azinovi\'c, Li, Kaplanyan, and
  Nie{\ss}ner}{Azinovi\'c et~al\mbox{.}}{2019}]%
        {azinovic2019inverse}
\bibfield{author}{\bibinfo{person}{Dejan Azinovi\'c}, \bibinfo{person}{Tzu-Mao
  Li}, \bibinfo{person}{Anton Kaplanyan}, {and} \bibinfo{person}{Matthias
  Nie{\ss}ner}.} \bibinfo{year}{2019}\natexlab{}.
\newblock \showarticletitle{Inverse Path Tracing for Joint Material and
  Lighting Estimation}. In \bibinfo{booktitle}{\emph{Proc. Computer Vision and
  Pattern Recognition (CVPR), IEEE}}.
\newblock


\bibitem[\protect\citeauthoryear{Bell, Upchurch, Snavely, and Bala}{Bell
  et~al\mbox{.}}{2013}]%
        {bell2013opensurfaces}
\bibfield{author}{\bibinfo{person}{Sean Bell}, \bibinfo{person}{Paul Upchurch},
  \bibinfo{person}{Noah Snavely}, {and} \bibinfo{person}{Kavita Bala}.}
  \bibinfo{year}{2013}\natexlab{}.
\newblock \showarticletitle{OpenSurfaces: A richly annotated catalog of surface
  appearance}.
\newblock \bibinfo{journal}{\emph{ACM Transactions on graphics (TOG)}}
  \bibinfo{volume}{32}, \bibinfo{number}{4} (\bibinfo{year}{2013}),
  \bibinfo{pages}{1--17}.
\newblock


\bibitem[\protect\citeauthoryear{Bell, Upchurch, Snavely, and Bala}{Bell
  et~al\mbox{.}}{2015}]%
        {bell15minc}
\bibfield{author}{\bibinfo{person}{Sean Bell}, \bibinfo{person}{Paul Upchurch},
  \bibinfo{person}{Noah Snavely}, {and} \bibinfo{person}{Kavita Bala}.}
  \bibinfo{year}{2015}\natexlab{}.
\newblock \showarticletitle{Material Recognition in the Wild with the Materials
  in Context Database}.
\newblock \bibinfo{journal}{\emph{Computer Vision and Pattern Recognition
  (CVPR)}} (\bibinfo{year}{2015}).
\newblock


\bibitem[\protect\citeauthoryear{Belongie, Carson, Greenspan, and
  Malik}{Belongie et~al\mbox{.}}{1998}]%
        {belongie1998color}
\bibfield{author}{\bibinfo{person}{Serge Belongie}, \bibinfo{person}{Chad
  Carson}, \bibinfo{person}{Hayit Greenspan}, {and} \bibinfo{person}{Jitendra
  Malik}.} \bibinfo{year}{1998}\natexlab{}.
\newblock \showarticletitle{Color-and texture-based image segmentation using EM
  and its application to content-based image retrieval}. In
  \bibinfo{booktitle}{\emph{Sixth International Conference on Computer Vision
  (IEEE Cat. No. 98CH36271)}}. IEEE, \bibinfo{pages}{675--682}.
\newblock


\bibitem[\protect\citeauthoryear{Bousseau, Paris, and Durand}{Bousseau
  et~al\mbox{.}}{2009}]%
        {Bousseau2009}
\bibfield{author}{\bibinfo{person}{Adrien Bousseau}, \bibinfo{person}{Sylvain
  Paris}, {and} \bibinfo{person}{Fr\'edo Durand}.}
  \bibinfo{year}{2009}\natexlab{}.
\newblock \showarticletitle{User Assisted Intrinsic Images}.
\newblock \bibinfo{journal}{\emph{ACM Transactions on Graphics (Proceedings of
  SIGGRAPH Asia 2009)}} \bibinfo{volume}{28}, \bibinfo{number}{5}
  (\bibinfo{year}{2009}).
\newblock


\bibitem[\protect\citeauthoryear{Caesar, Uijlings, and Ferrari}{Caesar
  et~al\mbox{.}}{2018}]%
        {caesar2018coco}
\bibfield{author}{\bibinfo{person}{Holger Caesar}, \bibinfo{person}{Jasper
  Uijlings}, {and} \bibinfo{person}{Vittorio Ferrari}.}
  \bibinfo{year}{2018}\natexlab{}.
\newblock \showarticletitle{Coco-stuff: Thing and stuff classes in context}. In
  \bibinfo{booktitle}{\emph{Proceedings of the IEEE conference on computer
  vision and pattern recognition}}. \bibinfo{pages}{1209--1218}.
\newblock


\bibitem[\protect\citeauthoryear{Cao, Cholakkal, Anwer, Khan, Pang, and
  Shao}{Cao et~al\mbox{.}}{2020}]%
        {cao2020d2det}
\bibfield{author}{\bibinfo{person}{Jiale Cao}, \bibinfo{person}{Hisham
  Cholakkal}, \bibinfo{person}{Rao~Muhammad Anwer},
  \bibinfo{person}{Fahad~Shahbaz Khan}, \bibinfo{person}{Yanwei Pang}, {and}
  \bibinfo{person}{Ling Shao}.} \bibinfo{year}{2020}\natexlab{}.
\newblock \showarticletitle{D2det: Towards high quality object detection and
  instance segmentation}. In \bibinfo{booktitle}{\emph{Proceedings of the
  IEEE/CVF conference on computer vision and pattern recognition}}.
  \bibinfo{pages}{11485--11494}.
\newblock


\bibitem[\protect\citeauthoryear{Cao, Li, Zhang, and Van~Gool}{Cao
  et~al\mbox{.}}{2021}]%
        {cao2021video}
\bibfield{author}{\bibinfo{person}{Jiezhang Cao}, \bibinfo{person}{Yawei Li},
  \bibinfo{person}{Kai Zhang}, {and} \bibinfo{person}{Luc Van~Gool}.}
  \bibinfo{year}{2021}\natexlab{}.
\newblock \showarticletitle{Video super-resolution transformer}.
\newblock \bibinfo{journal}{\emph{arXiv preprint arXiv:2106.06847}}
  (\bibinfo{year}{2021}).
\newblock


\bibitem[\protect\citeauthoryear{Caron, Touvron, Misra, J{\'e}gou, Mairal,
  Bojanowski, and Joulin}{Caron et~al\mbox{.}}{2021}]%
        {caron2021emerging}
\bibfield{author}{\bibinfo{person}{Mathilde Caron}, \bibinfo{person}{Hugo
  Touvron}, \bibinfo{person}{Ishan Misra}, \bibinfo{person}{Herv{\'e}
  J{\'e}gou}, \bibinfo{person}{Julien Mairal}, \bibinfo{person}{Piotr
  Bojanowski}, {and} \bibinfo{person}{Armand Joulin}.}
  \bibinfo{year}{2021}\natexlab{}.
\newblock \showarticletitle{Emerging properties in self-supervised vision
  transformers}. In \bibinfo{booktitle}{\emph{Proceedings of the IEEE/CVF
  International Conference on Computer Vision}}. \bibinfo{pages}{9650--9660}.
\newblock


\bibitem[\protect\citeauthoryear{Chen, Huang, Xia, and Zhang}{Chen
  et~al\mbox{.}}{2020}]%
        {chen2020nonlocal}
\bibfield{author}{\bibinfo{person}{Bingling Chen}, \bibinfo{person}{Yan Huang},
  \bibinfo{person}{Qiaoqiao Xia}, {and} \bibinfo{person}{Qinglin Zhang}.}
  \bibinfo{year}{2020}\natexlab{}.
\newblock \showarticletitle{Nonlocal spatial attention module for image
  classification}.
\newblock \bibinfo{journal}{\emph{International Journal of Advanced Robotic
  Systems}} \bibinfo{volume}{17}, \bibinfo{number}{5} (\bibinfo{year}{2020}),
  \bibinfo{pages}{1729881420938927}.
\newblock


\bibitem[\protect\citeauthoryear{Chen, Papandreou, Kokkinos, Murphy, and
  Yuille}{Chen et~al\mbox{.}}{2017}]%
        {chen2017deeplab}
\bibfield{author}{\bibinfo{person}{Liang-Chieh Chen}, \bibinfo{person}{George
  Papandreou}, \bibinfo{person}{Iasonas Kokkinos}, \bibinfo{person}{Kevin
  Murphy}, {and} \bibinfo{person}{Alan~L Yuille}.}
  \bibinfo{year}{2017}\natexlab{}.
\newblock \showarticletitle{Deeplab: Semantic image segmentation with deep
  convolutional nets, atrous convolution, and fully connected crfs}.
\newblock \bibinfo{journal}{\emph{IEEE transactions on pattern analysis and
  machine intelligence}} \bibinfo{volume}{40}, \bibinfo{number}{4}
  (\bibinfo{year}{2017}), \bibinfo{pages}{834--848}.
\newblock


\bibitem[\protect\citeauthoryear{Chen, Li, and Tang}{Chen
  et~al\mbox{.}}{2013}]%
        {knn_matting}
\bibfield{author}{\bibinfo{person}{Qifeng Chen}, \bibinfo{person}{Dingzeyu Li},
  {and} \bibinfo{person}{Chi-Keung Tang}.} \bibinfo{year}{2013}\natexlab{}.
\newblock \showarticletitle{KNN matting}.
\newblock \bibinfo{journal}{\emph{Pattern Analysis and Machine Intelligence,
  IEEE Transactions on}} \bibinfo{volume}{35}, \bibinfo{number}{9}
  (\bibinfo{date}{Sept} \bibinfo{year}{2013}), \bibinfo{pages}{2175--2188}.
\newblock
\urldef\tempurl%
\url{https://doi.org/10.1109/TPAMI.2013.18}
\showDOI{\tempurl}


\bibitem[\protect\citeauthoryear{Chen, Wang, Zhou, and Dong}{Chen
  et~al\mbox{.}}{2022}]%
        {chen2022activating}
\bibfield{author}{\bibinfo{person}{Xiangyu Chen}, \bibinfo{person}{Xintao
  Wang}, \bibinfo{person}{Jiantao Zhou}, {and} \bibinfo{person}{Chao Dong}.}
  \bibinfo{year}{2022}\natexlab{}.
\newblock \showarticletitle{Activating More Pixels in Image Super-Resolution
  Transformer}.
\newblock \bibinfo{journal}{\emph{arXiv preprint arXiv:2205.04437}}
  (\bibinfo{year}{2022}).
\newblock


\bibitem[\protect\citeauthoryear{Cimpoi, Maji, and Vedaldi}{Cimpoi
  et~al\mbox{.}}{2014}]%
        {cimpoi2014deep}
\bibfield{author}{\bibinfo{person}{Mircea Cimpoi}, \bibinfo{person}{Subhransu
  Maji}, {and} \bibinfo{person}{Andrea Vedaldi}.}
  \bibinfo{year}{2014}\natexlab{}.
\newblock \showarticletitle{Deep convolutional filter banks for texture
  recognition and segmentation}.
\newblock \bibinfo{journal}{\emph{arXiv preprint arXiv:1411.6836}}
  (\bibinfo{year}{2014}).
\newblock


\bibitem[\protect\citeauthoryear{Collins, Goel, Deng, Luthra, Xu, Gundogdu,
  Zhang, Yago~Vicente, Dideriksen, Arora, Guillaumin, and Malik}{Collins
  et~al\mbox{.}}{2022}]%
        {collins2022abo}
\bibfield{author}{\bibinfo{person}{Jasmine Collins}, \bibinfo{person}{Shubham
  Goel}, \bibinfo{person}{Kenan Deng}, \bibinfo{person}{Achleshwar Luthra},
  \bibinfo{person}{Leon Xu}, \bibinfo{person}{Erhan Gundogdu},
  \bibinfo{person}{Xi Zhang}, \bibinfo{person}{Tomas~F Yago~Vicente},
  \bibinfo{person}{Thomas Dideriksen}, \bibinfo{person}{Himanshu Arora},
  \bibinfo{person}{Matthieu Guillaumin}, {and} \bibinfo{person}{Jitendra
  Malik}.} \bibinfo{year}{2022}\natexlab{}.
\newblock \showarticletitle{ABO: Dataset and Benchmarks for Real-World 3D
  Object Understanding}.
\newblock \bibinfo{journal}{\emph{CVPR}} (\bibinfo{year}{2022}).
\newblock


\bibitem[\protect\citeauthoryear{Community}{Community}{2018}]%
        {blender}
\bibfield{author}{\bibinfo{person}{Blender~Online Community}.}
  \bibinfo{year}{2018}\natexlab{}.
\newblock \bibinfo{title}{Blender - a 3D modelling and rendering package}.
\newblock
\newblock
\urldef\tempurl%
\url{http://www.blender.org}
\showURL{%
\tempurl}


\bibitem[\protect\citeauthoryear{Cordts, Omran, Ramos, Rehfeld, Enzweiler,
  Benenson, Franke, Roth, and Schiele}{Cordts et~al\mbox{.}}{2016}]%
        {cordts2016cityscapes}
\bibfield{author}{\bibinfo{person}{Marius Cordts}, \bibinfo{person}{Mohamed
  Omran}, \bibinfo{person}{Sebastian Ramos}, \bibinfo{person}{Timo Rehfeld},
  \bibinfo{person}{Markus Enzweiler}, \bibinfo{person}{Rodrigo Benenson},
  \bibinfo{person}{Uwe Franke}, \bibinfo{person}{Stefan Roth}, {and}
  \bibinfo{person}{Bernt Schiele}.} \bibinfo{year}{2016}\natexlab{}.
\newblock \showarticletitle{The cityscapes dataset for semantic urban scene
  understanding}. In \bibinfo{booktitle}{\emph{Proceedings of the IEEE
  conference on computer vision and pattern recognition}}.
  \bibinfo{pages}{3213--3223}.
\newblock


\bibitem[\protect\citeauthoryear{Deng and Manjunath}{Deng and
  Manjunath}{2001}]%
        {deng2001unsupervised}
\bibfield{author}{\bibinfo{person}{Yining Deng} {and}
  \bibinfo{person}{Bangalore~S Manjunath}.} \bibinfo{year}{2001}\natexlab{}.
\newblock \showarticletitle{Unsupervised segmentation of color-texture regions
  in images and video}.
\newblock \bibinfo{journal}{\emph{IEEE transactions on pattern analysis and
  machine intelligence}} \bibinfo{volume}{23}, \bibinfo{number}{8}
  (\bibinfo{year}{2001}), \bibinfo{pages}{800--810}.
\newblock


\bibitem[\protect\citeauthoryear{Deschaintre, Aittala, Durand, Drettakis, and
  Bousseau}{Deschaintre et~al\mbox{.}}{2018}]%
        {Deschaintre2018}
\bibfield{author}{\bibinfo{person}{Valentin Deschaintre},
  \bibinfo{person}{Miika Aittala}, \bibinfo{person}{Fredo Durand},
  \bibinfo{person}{George Drettakis}, {and} \bibinfo{person}{Adrien Bousseau}.}
  \bibinfo{year}{2018}\natexlab{}.
\newblock \showarticletitle{Single-image SVBRDF Capture with a Rendering-aware
  Deep Network}.
\newblock \bibinfo{journal}{\emph{ACM Trans. Graph.}} \bibinfo{volume}{37},
  \bibinfo{number}{4} (\bibinfo{year}{2018}), \bibinfo{pages}{128:1--128:15}.
\newblock


\bibitem[\protect\citeauthoryear{Deschaintre, Aittala, Durand, Drettakis, and
  Bousseau}{Deschaintre et~al\mbox{.}}{2019}]%
        {Deschaintre2019}
\bibfield{author}{\bibinfo{person}{Valentin Deschaintre},
  \bibinfo{person}{Miika Aittala}, \bibinfo{person}{Fr\'edo Durand},
  \bibinfo{person}{George Drettakis}, {and} \bibinfo{person}{Adrien Bousseau}.}
  \bibinfo{year}{2019}\natexlab{}.
\newblock \showarticletitle{Flexible SVBRDF Capture with a Multi-Image Deep
  Network}.
\newblock \bibinfo{journal}{\emph{Computer Graphics Forum}}
  \bibinfo{volume}{38}, \bibinfo{number}{4} (\bibinfo{year}{2019}).
\newblock


\bibitem[\protect\citeauthoryear{Deschaintre, Drettakis, and
  Bousseau}{Deschaintre et~al\mbox{.}}{2020}]%
        {Deschaintre2020}
\bibfield{author}{\bibinfo{person}{Valentin Deschaintre},
  \bibinfo{person}{George Drettakis}, {and} \bibinfo{person}{Adrien Bousseau}.}
  \bibinfo{year}{2020}\natexlab{}.
\newblock \showarticletitle{Guided Fine-Tuning for Large-Scale Material
  Transfer}.
\newblock \bibinfo{journal}{\emph{Computer Graphics Forum (Proceedings of the
  Eurographics Symposium on Rendering)}} \bibinfo{volume}{39},
  \bibinfo{number}{4} (\bibinfo{year}{2020}).
\newblock
\urldef\tempurl%
\url{http://www-sop.inria.fr/reves/Basilic/2020/DDB20}
\showURL{%
\tempurl}


\bibitem[\protect\citeauthoryear{Deschaintre, Lin, and Ghosh}{Deschaintre
  et~al\mbox{.}}{2021}]%
        {Deschaintre2021}
\bibfield{author}{\bibinfo{person}{Valentin Deschaintre},
  \bibinfo{person}{Yiming Lin}, {and} \bibinfo{person}{Abhijeet Ghosh}.}
  \bibinfo{year}{2021}\natexlab{}.
\newblock \showarticletitle{Deep polarization imaging for 3D shape and SVBRDF
  acquisition}. In \bibinfo{booktitle}{\emph{Proceedings of the IEEE/CVF
  Conference on Computer Vision and Pattern Recognition (CVPR)}}.
\newblock


\bibitem[\protect\citeauthoryear{Dosovitskiy, Beyer, Kolesnikov, Weissenborn,
  Zhai, Unterthiner, Dehghani, Minderer, Heigold, Gelly,
  et~al\mbox{.}}{Dosovitskiy et~al\mbox{.}}{2020}]%
        {dosovitskiy2020image}
\bibfield{author}{\bibinfo{person}{Alexey Dosovitskiy}, \bibinfo{person}{Lucas
  Beyer}, \bibinfo{person}{Alexander Kolesnikov}, \bibinfo{person}{Dirk
  Weissenborn}, \bibinfo{person}{Xiaohua Zhai}, \bibinfo{person}{Thomas
  Unterthiner}, \bibinfo{person}{Mostafa Dehghani}, \bibinfo{person}{Matthias
  Minderer}, \bibinfo{person}{Georg Heigold}, \bibinfo{person}{Sylvain Gelly},
  {et~al\mbox{.}}} \bibinfo{year}{2020}\natexlab{}.
\newblock \showarticletitle{An image is worth 16x16 words: Transformers for
  image recognition at scale}.
\newblock \bibinfo{journal}{\emph{arXiv preprint arXiv:2010.11929}}
  (\bibinfo{year}{2020}).
\newblock


\bibitem[\protect\citeauthoryear{Everingham, Eslami, Van~Gool, Williams, Winn,
  and Zisserman}{Everingham et~al\mbox{.}}{2015}]%
        {Everingham15}
\bibfield{author}{\bibinfo{person}{M. Everingham}, \bibinfo{person}{S.~M.~A.
  Eslami}, \bibinfo{person}{L. Van~Gool}, \bibinfo{person}{C.~K.~I. Williams},
  \bibinfo{person}{J. Winn}, {and} \bibinfo{person}{A. Zisserman}.}
  \bibinfo{year}{2015}\natexlab{}.
\newblock \showarticletitle{The Pascal Visual Object Classes Challenge: A
  Retrospective}.
\newblock \bibinfo{journal}{\emph{International Journal of Computer Vision}}
  \bibinfo{volume}{111}, \bibinfo{number}{1} (\bibinfo{date}{Jan.}
  \bibinfo{year}{2015}), \bibinfo{pages}{98--136}.
\newblock


\bibitem[\protect\citeauthoryear{Fogel and Sagi}{Fogel and Sagi}{1989}]%
        {fogel1989gabor}
\bibfield{author}{\bibinfo{person}{Itzhak Fogel} {and} \bibinfo{person}{Dov
  Sagi}.} \bibinfo{year}{1989}\natexlab{}.
\newblock \showarticletitle{Gabor filters as texture discriminator}.
\newblock \bibinfo{journal}{\emph{Biological cybernetics}}
  \bibinfo{volume}{61}, \bibinfo{number}{2} (\bibinfo{year}{1989}),
  \bibinfo{pages}{103--113}.
\newblock


\bibitem[\protect\citeauthoryear{Gao, Li, Dong, Peers, Xu, and Tong}{Gao
  et~al\mbox{.}}{2019}]%
        {Gao2019}
\bibfield{author}{\bibinfo{person}{Duan Gao}, \bibinfo{person}{Xiao Li},
  \bibinfo{person}{Yue Dong}, \bibinfo{person}{Pieter Peers},
  \bibinfo{person}{Kun Xu}, {and} \bibinfo{person}{Xin Tong}.}
  \bibinfo{year}{2019}\natexlab{}.
\newblock \showarticletitle{Deep inverse rendering for high-resolution SVBRDF
  estimation from an arbitrary number of images}.
\newblock \bibinfo{journal}{\emph{ACM Trans. Graph.}} \bibinfo{volume}{38},
  \bibinfo{number}{4} (\bibinfo{year}{2019}).
\newblock


\bibitem[\protect\citeauthoryear{Garces, Rodriguez-Pardo, Casas, and
  Lopez-Moreno}{Garces et~al\mbox{.}}{2022}]%
        {garces2022survey}
\bibfield{author}{\bibinfo{person}{Elena Garces}, \bibinfo{person}{Carlos
  Rodriguez-Pardo}, \bibinfo{person}{Dan Casas}, {and} \bibinfo{person}{Jorge
  Lopez-Moreno}.} \bibinfo{year}{2022}\natexlab{}.
\newblock \showarticletitle{A Survey on Intrinsic Images: Delving Deep into
  Lambert and Beyond}.
\newblock \bibinfo{journal}{\emph{International Journal of Computer Vision}}
  \bibinfo{volume}{130}, \bibinfo{number}{3} (\bibinfo{year}{2022}),
  \bibinfo{pages}{836--868}.
\newblock


\bibitem[\protect\citeauthoryear{Geiger, Lenz, and Urtasun}{Geiger
  et~al\mbox{.}}{2012}]%
        {geiger2012we}
\bibfield{author}{\bibinfo{person}{Andreas Geiger}, \bibinfo{person}{Philip
  Lenz}, {and} \bibinfo{person}{Raquel Urtasun}.}
  \bibinfo{year}{2012}\natexlab{}.
\newblock \showarticletitle{Are we ready for autonomous driving? the kitti
  vision benchmark suite}. In \bibinfo{booktitle}{\emph{2012 IEEE conference on
  computer vision and pattern recognition}}. IEEE, \bibinfo{pages}{3354--3361}.
\newblock


\bibitem[\protect\citeauthoryear{Germer, Uelwer, Conrad, and Harmeling}{Germer
  et~al\mbox{.}}{2020}]%
        {pymatting}
\bibfield{author}{\bibinfo{person}{Thomas Germer}, \bibinfo{person}{Tobias
  Uelwer}, \bibinfo{person}{Stefan Conrad}, {and} \bibinfo{person}{Stefan
  Harmeling}.} \bibinfo{year}{2020}\natexlab{}.
\newblock \showarticletitle{PyMatting: A Python Library for Alpha Matting}.
\newblock \bibinfo{journal}{\emph{Journal of Open Source Software}}
  \bibinfo{volume}{5}, \bibinfo{number}{54} (\bibinfo{year}{2020}),
  \bibinfo{pages}{2481}.
\newblock
\urldef\tempurl%
\url{https://doi.org/10.21105/joss.02481}
\showDOI{\tempurl}


\bibitem[\protect\citeauthoryear{Ghiasi, Gu, Cui, and Lin}{Ghiasi
  et~al\mbox{.}}{2021}]%
        {ghiasi2021open}
\bibfield{author}{\bibinfo{person}{Golnaz Ghiasi}, \bibinfo{person}{Xiuye Gu},
  \bibinfo{person}{Yin Cui}, {and} \bibinfo{person}{Tsung-Yi Lin}.}
  \bibinfo{year}{2021}\natexlab{}.
\newblock \showarticletitle{Open-vocabulary image segmentation}.
\newblock \bibinfo{journal}{\emph{arXiv preprint arXiv:2112.12143}}
  (\bibinfo{year}{2021}).
\newblock


\bibitem[\protect\citeauthoryear{Griffiths, Ritschel, and Philip}{Griffiths
  et~al\mbox{.}}{2022}]%
        {griffiths2022outcast}
\bibfield{author}{\bibinfo{person}{David Griffiths}, \bibinfo{person}{Tobias
  Ritschel}, {and} \bibinfo{person}{Julien Philip}.}
  \bibinfo{year}{2022}\natexlab{}.
\newblock \showarticletitle{OutCast: Single Image Relighting with Cast
  Shadows}.
\newblock \bibinfo{journal}{\emph{Computer Graphics Forum}}
  \bibinfo{volume}{43} (\bibinfo{year}{2022}).
\newblock


\bibitem[\protect\citeauthoryear{Guarnera, Guarnera, Ghosh, Denk, and
  Glencross}{Guarnera et~al\mbox{.}}{2016}]%
        {Guarnera2016}
\bibfield{author}{\bibinfo{person}{Dar'ya Guarnera},
  \bibinfo{person}{Giuseppe~Claudio Guarnera}, \bibinfo{person}{Abhijeet
  Ghosh}, \bibinfo{person}{Cornelia Denk}, {and} \bibinfo{person}{Mashhuda
  Glencross}.} \bibinfo{year}{2016}\natexlab{}.
\newblock \showarticletitle{BRDF Representation and Acquisition}.
\newblock \bibinfo{journal}{\emph{Computer Graphics Forum}}
  (\bibinfo{year}{2016}).
\newblock


\bibitem[\protect\citeauthoryear{Guo, Lai, Tao, Cai, Wang, Guo, and Yan}{Guo
  et~al\mbox{.}}{2021}]%
        {Guo2021}
\bibfield{author}{\bibinfo{person}{Jie Guo}, \bibinfo{person}{Shuichang Lai},
  \bibinfo{person}{Chengzhi Tao}, \bibinfo{person}{Yuelong Cai},
  \bibinfo{person}{Lei Wang}, \bibinfo{person}{Yanwen Guo}, {and}
  \bibinfo{person}{Ling-Qi Yan}.} \bibinfo{year}{2021}\natexlab{}.
\newblock \showarticletitle{Highlight-Aware Two-Stream Network for Single-Image
  SVBRDF Acquisition}.
\newblock \bibinfo{journal}{\emph{ACM Trans. Graph.}} \bibinfo{volume}{40},
  \bibinfo{number}{4}, Article \bibinfo{articleno}{123} (\bibinfo{date}{jul}
  \bibinfo{year}{2021}), \bibinfo{numpages}{14}~pages.
\newblock
\showISSN{0730-0301}
\urldef\tempurl%
\url{https://doi.org/10.1145/3450626.3459854}
\showDOI{\tempurl}


\bibitem[\protect\citeauthoryear{Guo, Smith, Ha\v{s}an, Sunkavalli, and
  Zhao}{Guo et~al\mbox{.}}{2020}]%
        {Guo2020}
\bibfield{author}{\bibinfo{person}{Yu Guo}, \bibinfo{person}{Cameron Smith},
  \bibinfo{person}{Milo\v{s} Ha\v{s}an}, \bibinfo{person}{Kalyan Sunkavalli},
  {and} \bibinfo{person}{Shuang Zhao}.} \bibinfo{year}{2020}\natexlab{}.
\newblock \showarticletitle{MaterialGAN: Reflectance Capture using a Generative
  SVBRDF Model}.
\newblock \bibinfo{journal}{\emph{ACM Trans. Graph.}} \bibinfo{volume}{39},
  \bibinfo{number}{6} (\bibinfo{year}{2020}), \bibinfo{pages}{254:1--254:13}.
\newblock


\bibitem[\protect\citeauthoryear{Haindl and Mikes}{Haindl and Mikes}{2008}]%
        {haindl2008texture}
\bibfield{author}{\bibinfo{person}{Michal Haindl} {and}
  \bibinfo{person}{Stanislav Mikes}.} \bibinfo{year}{2008}\natexlab{}.
\newblock \showarticletitle{Texture segmentation benchmark}. In
  \bibinfo{booktitle}{\emph{2008 19th International Conference on Pattern
  Recognition}}. IEEE, \bibinfo{pages}{1--4}.
\newblock


\bibitem[\protect\citeauthoryear{Hamilton, Zhang, Hariharan, Snavely, and
  Freeman}{Hamilton et~al\mbox{.}}{2022}]%
        {hamilton2022unsupervised}
\bibfield{author}{\bibinfo{person}{Mark Hamilton}, \bibinfo{person}{Zhoutong
  Zhang}, \bibinfo{person}{Bharath Hariharan}, \bibinfo{person}{Noah Snavely},
  {and} \bibinfo{person}{William~T Freeman}.} \bibinfo{year}{2022}\natexlab{}.
\newblock \showarticletitle{Unsupervised Semantic Segmentation by Distilling
  Feature Correspondences}.
\newblock \bibinfo{journal}{\emph{arXiv preprint arXiv:2203.08414}}
  (\bibinfo{year}{2022}).
\newblock


\bibitem[\protect\citeauthoryear{Haralick, Shanmugam, and Dinstein}{Haralick
  et~al\mbox{.}}{1973}]%
        {haralick1973textural}
\bibfield{author}{\bibinfo{person}{Robert~M Haralick},
  \bibinfo{person}{Karthikeyan Shanmugam}, {and} \bibinfo{person}{Its'~Hak
  Dinstein}.} \bibinfo{year}{1973}\natexlab{}.
\newblock \showarticletitle{Textural features for image classification}.
\newblock \bibinfo{journal}{\emph{IEEE Transactions on systems, man, and
  cybernetics}} \bibinfo{number}{6} (\bibinfo{year}{1973}),
  \bibinfo{pages}{610--621}.
\newblock


\bibitem[\protect\citeauthoryear{He, Gkioxari, Doll{\'a}r, and Girshick}{He
  et~al\mbox{.}}{2017}]%
        {he2017mask}
\bibfield{author}{\bibinfo{person}{Kaiming He}, \bibinfo{person}{Georgia
  Gkioxari}, \bibinfo{person}{Piotr Doll{\'a}r}, {and} \bibinfo{person}{Ross
  Girshick}.} \bibinfo{year}{2017}\natexlab{}.
\newblock \showarticletitle{Mask r-cnn}. In
  \bibinfo{booktitle}{\emph{Proceedings of the IEEE international conference on
  computer vision}}. \bibinfo{pages}{2961--2969}.
\newblock


\bibitem[\protect\citeauthoryear{Henzler, Deschaintre, Mitra, and
  Ritschel}{Henzler et~al\mbox{.}}{2021}]%
        {Henzler2021}
\bibfield{author}{\bibinfo{person}{Philipp Henzler}, \bibinfo{person}{Valentin
  Deschaintre}, \bibinfo{person}{Niloy~J. Mitra}, {and} \bibinfo{person}{Tobias
  Ritschel}.} \bibinfo{year}{2021}\natexlab{}.
\newblock \showarticletitle{Generative Modelling of BRDF Textures from Flash
  Images}.
\newblock \bibinfo{journal}{\emph{ACM Trans. Graph.}} \bibinfo{volume}{40},
  \bibinfo{number}{6}, Article \bibinfo{articleno}{284} (\bibinfo{date}{dec}
  \bibinfo{year}{2021}), \bibinfo{numpages}{13}~pages.
\newblock


\bibitem[\protect\citeauthoryear{Hu, Shen, and Sun}{Hu et~al\mbox{.}}{2018}]%
        {hu2018squeeze}
\bibfield{author}{\bibinfo{person}{Jie Hu}, \bibinfo{person}{Li Shen}, {and}
  \bibinfo{person}{Gang Sun}.} \bibinfo{year}{2018}\natexlab{}.
\newblock \showarticletitle{Squeeze-and-excitation networks}. In
  \bibinfo{booktitle}{\emph{Proceedings of the IEEE conference on computer
  vision and pattern recognition}}. \bibinfo{pages}{7132--7141}.
\newblock


\bibitem[\protect\citeauthoryear{Hu, Hašan, Guerrero, Rushmeier, and
  Deschaintre}{Hu et~al\mbox{.}}{2022a}]%
        {hu2022control}
\bibfield{author}{\bibinfo{person}{Yiwei Hu}, \bibinfo{person}{Miloš Hašan},
  \bibinfo{person}{Paul Guerrero}, \bibinfo{person}{Holly Rushmeier}, {and}
  \bibinfo{person}{Valentin Deschaintre}.} \bibinfo{year}{2022}\natexlab{a}.
\newblock \showarticletitle{{Controlling Material Appearance by Examples}}.
\newblock \bibinfo{journal}{\emph{Computer Graphics Forum}}
  (\bibinfo{year}{2022}).
\newblock
\showISSN{1467-8659}
\urldef\tempurl%
\url{https://doi.org/10.1111/cgf.14591}
\showDOI{\tempurl}


\bibitem[\protect\citeauthoryear{Hu, He, Deschaintre, Dorsey, and Rushmeier}{Hu
  et~al\mbox{.}}{2022b}]%
        {hu2022}
\bibfield{author}{\bibinfo{person}{Yiwei Hu}, \bibinfo{person}{Chengan He},
  \bibinfo{person}{Valentin Deschaintre}, \bibinfo{person}{Julie Dorsey}, {and}
  \bibinfo{person}{Holly Rushmeier}.} \bibinfo{year}{2022}\natexlab{b}.
\newblock \showarticletitle{An Inverse Procedural Modeling Pipeline for SVBRDF
  Maps}.
\newblock \bibinfo{journal}{\emph{ACM Trans. Graph.}} \bibinfo{volume}{41},
  \bibinfo{number}{2}, Article \bibinfo{articleno}{18} (\bibinfo{date}{jan}
  \bibinfo{year}{2022}), \bibinfo{numpages}{17}~pages.
\newblock
\showISSN{0730-0301}
\urldef\tempurl%
\url{https://doi.org/10.1145/3502431}
\showDOI{\tempurl}


\bibitem[\protect\citeauthoryear{Karimi~Dastjerdi, Hold-Geoffroy, Eisenmann,
  Khodadadeh, and Lalonde}{Karimi~Dastjerdi et~al\mbox{.}}{2022}]%
        {karimi2022ImmerseGAN}
\bibfield{author}{\bibinfo{person}{Mohammad~Reza Karimi~Dastjerdi},
  \bibinfo{person}{Yannick Hold-Geoffroy}, \bibinfo{person}{Jonathan
  Eisenmann}, \bibinfo{person}{Siavash Khodadadeh}, {and}
  \bibinfo{person}{Jean-Fran{\c{c}}ois Lalonde}.}
  \bibinfo{year}{2022}\natexlab{}.
\newblock \showarticletitle{Guided Co-Modulated GAN for 360 degree Field of
  View Extrapolation}.
\newblock \bibinfo{journal}{\emph{International Conference on 3D Vision (3DV)}}
  (\bibinfo{year}{2022}).
\newblock


\bibitem[\protect\citeauthoryear{Khan, Reinhard, Fleming, and
  B\"{u}lthoff}{Khan et~al\mbox{.}}{2006}]%
        {10.1145/1141911.1141937}
\bibfield{author}{\bibinfo{person}{Erum~Arif Khan}, \bibinfo{person}{Erik
  Reinhard}, \bibinfo{person}{Roland~W. Fleming}, {and}
  \bibinfo{person}{Heinrich~H. B\"{u}lthoff}.} \bibinfo{year}{2006}\natexlab{}.
\newblock \showarticletitle{Image-Based Material Editing}.
\newblock \bibinfo{journal}{\emph{ACM Trans. Graph.}} \bibinfo{volume}{25},
  \bibinfo{number}{3} (\bibinfo{date}{jul} \bibinfo{year}{2006}),
  \bibinfo{pages}{654–663}.
\newblock
\showISSN{0730-0301}
\urldef\tempurl%
\url{https://doi.org/10.1145/1141911.1141937}
\showDOI{\tempurl}


\bibitem[\protect\citeauthoryear{Lawrence, Ben-Artzi, DeCoro, Matusik, Pfister,
  Ramamoorthi, and Rusinkiewicz}{Lawrence et~al\mbox{.}}{2006}]%
        {Lawrence2006}
\bibfield{author}{\bibinfo{person}{Jason Lawrence}, \bibinfo{person}{Aner
  Ben-Artzi}, \bibinfo{person}{Christopher DeCoro}, \bibinfo{person}{Wojciech
  Matusik}, \bibinfo{person}{Hanspeter Pfister}, \bibinfo{person}{Ravi
  Ramamoorthi}, {and} \bibinfo{person}{Szymon Rusinkiewicz}.}
  \bibinfo{year}{2006}\natexlab{}.
\newblock \showarticletitle{{Inverse Shade Trees for Non-Parametric Material
  Representation and Editing}}.
\newblock \bibinfo{journal}{\emph{ACM Transactions on Graphics (Proc.
  SIGGRAPH)}} \bibinfo{volume}{25}, \bibinfo{number}{3} (\bibinfo{date}{July}
  \bibinfo{year}{2006}).
\newblock


\bibitem[\protect\citeauthoryear{Lepage and Lawrence}{Lepage and
  Lawrence}{2011}]%
        {Lepage11}
\bibfield{author}{\bibinfo{person}{Daniel Lepage} {and} \bibinfo{person}{Jason
  Lawrence}.} \bibinfo{year}{2011}\natexlab{}.
\newblock \showarticletitle{Material Matting}.
\newblock \bibinfo{journal}{\emph{ACM Trans. Graph.}} \bibinfo{volume}{30},
  \bibinfo{number}{6} (\bibinfo{date}{Dec.} \bibinfo{year}{2011}),
  \bibinfo{pages}{1–10}.
\newblock
\showISSN{0730-0301}
\urldef\tempurl%
\url{https://doi.org/10.1145/2070781.2024178}
\showDOI{\tempurl}


\bibitem[\protect\citeauthoryear{Leung and Malik}{Leung and Malik}{2001}]%
        {leung2001representing}
\bibfield{author}{\bibinfo{person}{Thomas Leung} {and}
  \bibinfo{person}{Jitendra Malik}.} \bibinfo{year}{2001}\natexlab{}.
\newblock \showarticletitle{Representing and recognizing the visual appearance
  of materials using three-dimensional textons}.
\newblock \bibinfo{journal}{\emph{International journal of computer vision}}
  \bibinfo{volume}{43}, \bibinfo{number}{1} (\bibinfo{year}{2001}),
  \bibinfo{pages}{29--44}.
\newblock


\bibitem[\protect\citeauthoryear{Li, Shafiei, Ramamoorthi, Sunkavalli, and
  Chandraker}{Li et~al\mbox{.}}{2020}]%
        {li2020inverse}
\bibfield{author}{\bibinfo{person}{Zhengqin Li}, \bibinfo{person}{Mohammad
  Shafiei}, \bibinfo{person}{Ravi Ramamoorthi}, \bibinfo{person}{Kalyan
  Sunkavalli}, {and} \bibinfo{person}{Manmohan Chandraker}.}
  \bibinfo{year}{2020}\natexlab{}.
\newblock \showarticletitle{Inverse rendering for complex indoor scenes: Shape,
  spatially-varying lighting and svbrdf from a single image}. In
  \bibinfo{booktitle}{\emph{Proceedings of the IEEE/CVF Conference on Computer
  Vision and Pattern Recognition}}. \bibinfo{pages}{2475--2484}.
\newblock


\bibitem[\protect\citeauthoryear{Li, Sunkavalli, and Chandraker}{Li
  et~al\mbox{.}}{2018a}]%
        {Li2018}
\bibfield{author}{\bibinfo{person}{Zhengqin Li}, \bibinfo{person}{Kalyan
  Sunkavalli}, {and} \bibinfo{person}{Manmohan Chandraker}.}
  \bibinfo{year}{2018}\natexlab{a}.
\newblock \showarticletitle{Materials for Masses: {SVBRDF} Acquisition with a
  Single Mobile Phone Image}. In \bibinfo{booktitle}{\emph{Computer Vision -
  {ECCV} 2018 - 15th European Conference, Munich, Germany, September 8-14,
  2018, Proceedings, Part {III}}} \emph{(\bibinfo{series}{Lecture Notes in
  Computer Science}, Vol.~\bibinfo{volume}{11207})}. \bibinfo{pages}{74--90}.
\newblock


\bibitem[\protect\citeauthoryear{Li, Xu, Ramamoorthi, Sunkavalli, and
  Chandraker}{Li et~al\mbox{.}}{2018b}]%
        {li2018_3D}
\bibfield{author}{\bibinfo{person}{Zhengqin Li}, \bibinfo{person}{Zexiang Xu},
  \bibinfo{person}{Ravi Ramamoorthi}, \bibinfo{person}{Kalyan Sunkavalli},
  {and} \bibinfo{person}{Manmohan Chandraker}.}
  \bibinfo{year}{2018}\natexlab{b}.
\newblock \showarticletitle{Learning to reconstruct shape and spatially-varying
  reflectance from a single image}. In \bibinfo{booktitle}{\emph{SIGGRAPH Asia
  2018 Technical Papers}}. ACM, \bibinfo{pages}{269}.
\newblock


\bibitem[\protect\citeauthoryear{Lin, Maire, Belongie, Hays, Perona, Ramanan,
  Doll{\'a}r, and Zitnick}{Lin et~al\mbox{.}}{2014}]%
        {lin2014microsoft}
\bibfield{author}{\bibinfo{person}{Tsung-Yi Lin}, \bibinfo{person}{Michael
  Maire}, \bibinfo{person}{Serge Belongie}, \bibinfo{person}{James Hays},
  \bibinfo{person}{Pietro Perona}, \bibinfo{person}{Deva Ramanan},
  \bibinfo{person}{Piotr Doll{\'a}r}, {and} \bibinfo{person}{C~Lawrence
  Zitnick}.} \bibinfo{year}{2014}\natexlab{}.
\newblock \showarticletitle{Microsoft coco: Common objects in context}. In
  \bibinfo{booktitle}{\emph{European conference on computer vision}}. Springer,
  \bibinfo{pages}{740--755}.
\newblock


\bibitem[\protect\citeauthoryear{Liu, Sharan, Adelson, and Rosenholtz}{Liu
  et~al\mbox{.}}{2010}]%
        {liu2010exploring}
\bibfield{author}{\bibinfo{person}{Ce Liu}, \bibinfo{person}{Lavanya Sharan},
  \bibinfo{person}{Edward~H Adelson}, {and} \bibinfo{person}{Ruth Rosenholtz}.}
  \bibinfo{year}{2010}\natexlab{}.
\newblock \showarticletitle{Exploring features in a bayesian framework for
  material recognition}. In \bibinfo{booktitle}{\emph{2010 ieee computer
  society conference on computer vision and pattern recognition}}. IEEE,
  \bibinfo{pages}{239--246}.
\newblock


\bibitem[\protect\citeauthoryear{Long, Shelhamer, and Darrell}{Long
  et~al\mbox{.}}{2015}]%
        {long2015fully}
\bibfield{author}{\bibinfo{person}{Jonathan Long}, \bibinfo{person}{Evan
  Shelhamer}, {and} \bibinfo{person}{Trevor Darrell}.}
  \bibinfo{year}{2015}\natexlab{}.
\newblock \showarticletitle{Fully convolutional networks for semantic
  segmentation}. In \bibinfo{booktitle}{\emph{Proceedings of the IEEE
  conference on computer vision and pattern recognition}}.
  \bibinfo{pages}{3431--3440}.
\newblock


\bibitem[\protect\citeauthoryear{Lu, Liu, Li, and Zhang}{Lu
  et~al\mbox{.}}{2021}]%
        {lu2021efficient}
\bibfield{author}{\bibinfo{person}{Zhisheng Lu}, \bibinfo{person}{Hong Liu},
  \bibinfo{person}{Juncheng Li}, {and} \bibinfo{person}{Linlin Zhang}.}
  \bibinfo{year}{2021}\natexlab{}.
\newblock \showarticletitle{Efficient transformer for single image
  super-resolution}.
\newblock \bibinfo{journal}{\emph{arXiv preprint arXiv:2108.11084}}
  (\bibinfo{year}{2021}).
\newblock


\bibitem[\protect\citeauthoryear{Malpica, Ortu{\~n}o, and Santos}{Malpica
  et~al\mbox{.}}{2003}]%
        {malpica2003multichannel}
\bibfield{author}{\bibinfo{person}{Norberto Malpica}, \bibinfo{person}{Juan~E
  Ortu{\~n}o}, {and} \bibinfo{person}{Andres Santos}.}
  \bibinfo{year}{2003}\natexlab{}.
\newblock \showarticletitle{A multichannel watershed-based algorithm for
  supervised texture segmentation}.
\newblock \bibinfo{journal}{\emph{Pattern Recognition Letters}}
  \bibinfo{volume}{24}, \bibinfo{number}{9-10} (\bibinfo{year}{2003}),
  \bibinfo{pages}{1545--1554}.
\newblock


\bibitem[\protect\citeauthoryear{Murmann, Gharbi, Aittala, and Durand}{Murmann
  et~al\mbox{.}}{2019}]%
        {murmann2019dataset}
\bibfield{author}{\bibinfo{person}{Lukas Murmann}, \bibinfo{person}{Michael
  Gharbi}, \bibinfo{person}{Miika Aittala}, {and} \bibinfo{person}{Fredo
  Durand}.} \bibinfo{year}{2019}\natexlab{}.
\newblock \showarticletitle{A dataset of multi-illumination images in the
  wild}. In \bibinfo{booktitle}{\emph{Proceedings of the IEEE/CVF International
  Conference on Computer Vision}}. \bibinfo{pages}{4080--4089}.
\newblock


\bibitem[\protect\citeauthoryear{Nicolet, Philip, and Drettakis}{Nicolet
  et~al\mbox{.}}{2020}]%
        {10.1145/3384382.3384523}
\bibfield{author}{\bibinfo{person}{Baptiste Nicolet}, \bibinfo{person}{Julien
  Philip}, {and} \bibinfo{person}{George Drettakis}.}
  \bibinfo{year}{2020}\natexlab{}.
\newblock \showarticletitle{Repurposing a Relighting Network for Realistic
  Compositions of Captured Scenes}. In \bibinfo{booktitle}{\emph{Symposium on
  Interactive 3D Graphics and Games}} (San Francisco, CA, USA)
  \emph{(\bibinfo{series}{I3D '20})}. \bibinfo{publisher}{Association for
  Computing Machinery}, \bibinfo{address}{New York, NY, USA}, Article
  \bibinfo{articleno}{4}, \bibinfo{numpages}{9}~pages.
\newblock
\showISBNx{9781450375894}
\urldef\tempurl%
\url{https://doi.org/10.1145/3384382.3384523}
\showDOI{\tempurl}


\bibitem[\protect\citeauthoryear{Nimier-David, Dong, Jakob, and
  Kaplanyan}{Nimier-David et~al\mbox{.}}{2021}]%
        {nimierdavid2021material}
\bibfield{author}{\bibinfo{person}{Merlin Nimier-David}, \bibinfo{person}{Zhao
  Dong}, \bibinfo{person}{Wenzel Jakob}, {and} \bibinfo{person}{Anton
  Kaplanyan}.} \bibinfo{year}{2021}\natexlab{}.
\newblock \showarticletitle{{Material and Lighting Reconstruction for Complex
  Indoor Scenes with Texture-space Differentiable Rendering}}. In
  \bibinfo{booktitle}{\emph{Eurographics Symposium on Rendering - DL-only
  Track}}, \bibfield{editor}{\bibinfo{person}{Adrien Bousseau} {and}
  \bibinfo{person}{Morgan McGuire}} (Eds.). \bibinfo{publisher}{The
  Eurographics Association}.
\newblock
\showISBNx{978-3-03868-157-1}
\showISSN{1727-3463}
\urldef\tempurl%
\url{https://doi.org/10.2312/sr.20211292}
\showDOI{\tempurl}


\bibitem[\protect\citeauthoryear{Peebles and Xie}{Peebles and Xie}{2022}]%
        {peebles2022scalable}
\bibfield{author}{\bibinfo{person}{William Peebles} {and}
  \bibinfo{person}{Saining Xie}.} \bibinfo{year}{2022}\natexlab{}.
\newblock \showarticletitle{Scalable Diffusion Models with Transformers}.
\newblock \bibinfo{journal}{\emph{arXiv preprint arXiv:2212.09748}}
  (\bibinfo{year}{2022}).
\newblock


\bibitem[\protect\citeauthoryear{Pellacini and Lawrence}{Pellacini and
  Lawrence}{2007}]%
        {Pellacini07}
\bibfield{author}{\bibinfo{person}{Fabio Pellacini} {and}
  \bibinfo{person}{Jason Lawrence}.} \bibinfo{year}{2007}\natexlab{}.
\newblock \showarticletitle{AppWand: Editing Measured Materials Using
  Appearance-Driven Optimization}.
\newblock \bibinfo{journal}{\emph{ACM Trans. Graph.}} \bibinfo{volume}{26},
  \bibinfo{number}{3} (\bibinfo{date}{jul} \bibinfo{year}{2007}),
  \bibinfo{pages}{54–es}.
\newblock
\showISSN{0730-0301}
\urldef\tempurl%
\url{https://doi.org/10.1145/1276377.1276444}
\showDOI{\tempurl}


\bibitem[\protect\citeauthoryear{Philip, Gharbi, Zhou, Efros, and
  Drettakis}{Philip et~al\mbox{.}}{2019}]%
        {Philip19}
\bibfield{author}{\bibinfo{person}{Julien Philip}, \bibinfo{person}{Micha\"{e}l
  Gharbi}, \bibinfo{person}{Tinghui Zhou}, \bibinfo{person}{Alexei~A. Efros},
  {and} \bibinfo{person}{George Drettakis}.} \bibinfo{year}{2019}\natexlab{}.
\newblock \showarticletitle{Multi-View Relighting Using a Geometry-Aware
  Network}.
\newblock \bibinfo{journal}{\emph{ACM Trans. Graph.}} \bibinfo{volume}{38},
  \bibinfo{number}{4}, Article \bibinfo{articleno}{78} (\bibinfo{date}{jul}
  \bibinfo{year}{2019}), \bibinfo{numpages}{14}~pages.
\newblock
\showISSN{0730-0301}
\urldef\tempurl%
\url{https://doi.org/10.1145/3306346.3323013}
\showDOI{\tempurl}


\bibitem[\protect\citeauthoryear{Philip, Morgenthaler, Gharbi, and
  Drettakis}{Philip et~al\mbox{.}}{2021}]%
        {Philip21}
\bibfield{author}{\bibinfo{person}{Julien Philip}, \bibinfo{person}{S\'ebastien
  Morgenthaler}, \bibinfo{person}{Micha{\"e}l Gharbi}, {and}
  \bibinfo{person}{George Drettakis}.} \bibinfo{year}{2021}\natexlab{}.
\newblock \showarticletitle{Free-viewpoint Indoor Neural Relighting from
  Multi-view Stereo}.
\newblock \bibinfo{journal}{\emph{ACM Transactions on Graphics}}
  (\bibinfo{year}{2021}).
\newblock
\urldef\tempurl%
\url{http://www-sop.inria.fr/reves/Basilic/2021/PMGD21}
\showURL{%
\tempurl}


\bibitem[\protect\citeauthoryear{Randen and Husoy}{Randen and Husoy}{1999}]%
        {randen1999filtering}
\bibfield{author}{\bibinfo{person}{Trygve Randen} {and}
  \bibinfo{person}{John~Hakon Husoy}.} \bibinfo{year}{1999}\natexlab{}.
\newblock \showarticletitle{Filtering for texture classification: A comparative
  study}.
\newblock \bibinfo{journal}{\emph{IEEE Transactions on pattern analysis and
  machine intelligence}} \bibinfo{volume}{21}, \bibinfo{number}{4}
  (\bibinfo{year}{1999}), \bibinfo{pages}{291--310}.
\newblock


\bibitem[\protect\citeauthoryear{Ranftl, Bochkovskiy, and Koltun}{Ranftl
  et~al\mbox{.}}{2021}]%
        {DPT}
\bibfield{author}{\bibinfo{person}{Ren\'e Ranftl}, \bibinfo{person}{Alexey
  Bochkovskiy}, {and} \bibinfo{person}{Vladlen Koltun}.}
  \bibinfo{year}{2021}\natexlab{}.
\newblock \showarticletitle{Vision Transformers for Dense Prediction}. In
  \bibinfo{booktitle}{\emph{Proceedings of the IEEE/CVF International
  Conference on Computer Vision (ICCV)}}. \bibinfo{pages}{12179--12188}.
\newblock


\bibitem[\protect\citeauthoryear{Ranftl, Lasinger, Hafner, Schindler, and
  Koltun}{Ranftl et~al\mbox{.}}{2022}]%
        {Ranftl2022}
\bibfield{author}{\bibinfo{person}{Ren\'{e} Ranftl}, \bibinfo{person}{Katrin
  Lasinger}, \bibinfo{person}{David Hafner}, \bibinfo{person}{Konrad
  Schindler}, {and} \bibinfo{person}{Vladlen Koltun}.}
  \bibinfo{year}{2022}\natexlab{}.
\newblock \showarticletitle{Towards Robust Monocular Depth Estimation: Mixing
  Datasets for Zero-Shot Cross-Dataset Transfer}.
\newblock \bibinfo{journal}{\emph{IEEE Transactions on Pattern Analysis and
  Machine Intelligence}} \bibinfo{volume}{44}, \bibinfo{number}{3}
  (\bibinfo{year}{2022}).
\newblock


\bibitem[\protect\citeauthoryear{Reyes-Aldasoro and Bhalerao}{Reyes-Aldasoro
  and Bhalerao}{2006}]%
        {reyes2006bhattacharyya}
\bibfield{author}{\bibinfo{person}{Constantino~Carlos Reyes-Aldasoro} {and}
  \bibinfo{person}{Abhir Bhalerao}.} \bibinfo{year}{2006}\natexlab{}.
\newblock \showarticletitle{The Bhattacharyya space for feature selection and
  its application to texture segmentation}.
\newblock \bibinfo{journal}{\emph{Pattern Recognition}} \bibinfo{volume}{39},
  \bibinfo{number}{5} (\bibinfo{year}{2006}), \bibinfo{pages}{812--826}.
\newblock


\bibitem[\protect\citeauthoryear{Rombach, Blattmann, Lorenz, Esser, and
  Ommer}{Rombach et~al\mbox{.}}{2021}]%
        {rombach2021highresolution}
\bibfield{author}{\bibinfo{person}{Robin Rombach}, \bibinfo{person}{Andreas
  Blattmann}, \bibinfo{person}{Dominik Lorenz}, \bibinfo{person}{Patrick
  Esser}, {and} \bibinfo{person}{Björn Ommer}.}
  \bibinfo{year}{2021}\natexlab{}.
\newblock \bibinfo{title}{High-Resolution Image Synthesis with Latent Diffusion
  Models}.
\newblock
\newblock
\showeprint[arxiv]{2112.10752}~[cs.CV]


\bibitem[\protect\citeauthoryear{Ronneberger, Fischer, and Brox}{Ronneberger
  et~al\mbox{.}}{2015}]%
        {ronneberger2015u}
\bibfield{author}{\bibinfo{person}{Olaf Ronneberger}, \bibinfo{person}{Philipp
  Fischer}, {and} \bibinfo{person}{Thomas Brox}.}
  \bibinfo{year}{2015}\natexlab{}.
\newblock \showarticletitle{U-net: Convolutional networks for biomedical image
  segmentation}. In \bibinfo{booktitle}{\emph{International Conference on
  Medical image computing and computer-assisted intervention}}. Springer,
  \bibinfo{pages}{234--241}.
\newblock


\bibitem[\protect\citeauthoryear{Schwartz and Nishino}{Schwartz and
  Nishino}{2013}]%
        {Schwartz_2013_ICCV_Workshops}
\bibfield{author}{\bibinfo{person}{Gabriel Schwartz} {and} \bibinfo{person}{Ko
  Nishino}.} \bibinfo{year}{2013}\natexlab{}.
\newblock \showarticletitle{Visual Material Traits: Recognizing Per-Pixel
  Material Context}. In \bibinfo{booktitle}{\emph{Proceedings of the IEEE
  International Conference on Computer Vision (ICCV) Workshops}}.
\newblock


\bibitem[\protect\citeauthoryear{Schwartz and Nishino}{Schwartz and
  Nishino}{2016}]%
        {Schwartz2016MaterialRF}
\bibfield{author}{\bibinfo{person}{Gabriel Schwartz} {and} \bibinfo{person}{Ko
  Nishino}.} \bibinfo{year}{2016}\natexlab{}.
\newblock \showarticletitle{Material Recognition from Local Appearance in
  Global Context}.
\newblock \bibinfo{journal}{\emph{ArXiv}}  \bibinfo{volume}{abs/1611.09394}
  (\bibinfo{year}{2016}).
\newblock


\bibitem[\protect\citeauthoryear{Schwartz and Nishino}{Schwartz and
  Nishino}{2019}]%
        {schwartz2019recognizing}
\bibfield{author}{\bibinfo{person}{Gabriel Schwartz} {and} \bibinfo{person}{Ko
  Nishino}.} \bibinfo{year}{2019}\natexlab{}.
\newblock \showarticletitle{Recognizing material properties from images}.
\newblock \bibinfo{journal}{\emph{IEEE transactions on pattern analysis and
  machine intelligence}} \bibinfo{volume}{42}, \bibinfo{number}{8}
  (\bibinfo{year}{2019}), \bibinfo{pages}{1981--1995}.
\newblock


\bibitem[\protect\citeauthoryear{Schwartz and Nishino}{Schwartz and
  Nishino}{2020}]%
        {Schwartz20}
\bibfield{author}{\bibinfo{person}{Gabriel Schwartz} {and} \bibinfo{person}{Ko
  Nishino}.} \bibinfo{year}{2020}\natexlab{}.
\newblock \showarticletitle{Recognizing Material Properties from Images}.
\newblock \bibinfo{journal}{\emph{IEEE Transactions on Pattern Analysis and
  Machine Intelligence}} \bibinfo{volume}{42}, \bibinfo{number}{8}
  (\bibinfo{year}{2020}), \bibinfo{pages}{1981--1995}.
\newblock
\urldef\tempurl%
\url{https://doi.org/10.1109/TPAMI.2019.2907850}
\showDOI{\tempurl}


\bibitem[\protect\citeauthoryear{Tan and Le}{Tan and Le}{2019}]%
        {tan2019efficientnet}
\bibfield{author}{\bibinfo{person}{Mingxing Tan} {and} \bibinfo{person}{Quoc
  Le}.} \bibinfo{year}{2019}\natexlab{}.
\newblock \showarticletitle{Efficientnet: Rethinking model scaling for
  convolutional neural networks}. In \bibinfo{booktitle}{\emph{International
  conference on machine learning}}. PMLR, \bibinfo{pages}{6105--6114}.
\newblock


\bibitem[\protect\citeauthoryear{Tkachenko, Malyuk, Shevchenko, Holmanyuk, and
  Liubimov}{Tkachenko et~al\mbox{.}}{2021}]%
        {label_studio}
\bibfield{author}{\bibinfo{person}{Maxim Tkachenko}, \bibinfo{person}{Mikhail
  Malyuk}, \bibinfo{person}{Nikita Shevchenko}, \bibinfo{person}{Andrey
  Holmanyuk}, {and} \bibinfo{person}{Nikolai Liubimov}.}
  \bibinfo{year}{2020-2021}\natexlab{}.
\newblock \bibinfo{title}{{Label Studio}: Data labeling software}.
\newblock
\newblock
\urldef\tempurl%
\url{https://github.com/heartexlabs/label-studio}
\showURL{%
\tempurl}
\newblock
\shownote{Open source software available from
  https://github.com/heartexlabs/label-studio.}


\bibitem[\protect\citeauthoryear{Todorovic and Ahuja}{Todorovic and
  Ahuja}{2009}]%
        {todorovic2009texel}
\bibfield{author}{\bibinfo{person}{Sinisa Todorovic} {and}
  \bibinfo{person}{Narendra Ahuja}.} \bibinfo{year}{2009}\natexlab{}.
\newblock \showarticletitle{Texel-based texture segmentation}. In
  \bibinfo{booktitle}{\emph{2009 IEEE 12th International Conference on Computer
  Vision}}. IEEE, \bibinfo{pages}{841--848}.
\newblock


\bibitem[\protect\citeauthoryear{Upchurch and Niu}{Upchurch and Niu}{2022}]%
        {upchurch2022dense}
\bibfield{author}{\bibinfo{person}{Paul Upchurch} {and} \bibinfo{person}{Ransen
  Niu}.} \bibinfo{year}{2022}\natexlab{}.
\newblock \showarticletitle{A Dense Material Segmentation Dataset for Indoor
  and Outdoor Scene Parsing}. In \bibinfo{booktitle}{\emph{European Conference
  on Computer Vision}}. Springer, \bibinfo{pages}{450--466}.
\newblock


\bibitem[\protect\citeauthoryear{Vaswani, Shazeer, Parmar, Uszkoreit, Jones,
  Gomez, Kaiser, and Polosukhin}{Vaswani et~al\mbox{.}}{2017}]%
        {vaswani2017attention}
\bibfield{author}{\bibinfo{person}{Ashish Vaswani}, \bibinfo{person}{Noam
  Shazeer}, \bibinfo{person}{Niki Parmar}, \bibinfo{person}{Jakob Uszkoreit},
  \bibinfo{person}{Llion Jones}, \bibinfo{person}{Aidan~N Gomez},
  \bibinfo{person}{{\L}ukasz Kaiser}, {and} \bibinfo{person}{Illia
  Polosukhin}.} \bibinfo{year}{2017}\natexlab{}.
\newblock \showarticletitle{Attention is all you need}.
\newblock \bibinfo{journal}{\emph{Advances in neural information processing
  systems}}  \bibinfo{volume}{30} (\bibinfo{year}{2017}).
\newblock


\bibitem[\protect\citeauthoryear{Wang, Zhu, Adam, Yuille, and Chen}{Wang
  et~al\mbox{.}}{2021}]%
        {wang2021max}
\bibfield{author}{\bibinfo{person}{Huiyu Wang}, \bibinfo{person}{Yukun Zhu},
  \bibinfo{person}{Hartwig Adam}, \bibinfo{person}{Alan Yuille}, {and}
  \bibinfo{person}{Liang-Chieh Chen}.} \bibinfo{year}{2021}\natexlab{}.
\newblock \showarticletitle{Max-deeplab: End-to-end panoptic segmentation with
  mask transformers}. In \bibinfo{booktitle}{\emph{Proceedings of the IEEE/CVF
  conference on computer vision and pattern recognition}}.
  \bibinfo{pages}{5463--5474}.
\newblock


\bibitem[\protect\citeauthoryear{Wang, Zhu, Green, Adam, Yuille, and Chen}{Wang
  et~al\mbox{.}}{2020b}]%
        {wang2020axial}
\bibfield{author}{\bibinfo{person}{Huiyu Wang}, \bibinfo{person}{Yukun Zhu},
  \bibinfo{person}{Bradley Green}, \bibinfo{person}{Hartwig Adam},
  \bibinfo{person}{Alan Yuille}, {and} \bibinfo{person}{Liang-Chieh Chen}.}
  \bibinfo{year}{2020}\natexlab{b}.
\newblock \showarticletitle{Axial-deeplab: Stand-alone axial-attention for
  panoptic segmentation}. In \bibinfo{booktitle}{\emph{European Conference on
  Computer Vision}}. Springer, \bibinfo{pages}{108--126}.
\newblock


\bibitem[\protect\citeauthoryear{Wang, Zhang, Kong, Li, and Shen}{Wang
  et~al\mbox{.}}{2020a}]%
        {wang2020solov2}
\bibfield{author}{\bibinfo{person}{Xinlong Wang}, \bibinfo{person}{Rufeng
  Zhang}, \bibinfo{person}{Tao Kong}, \bibinfo{person}{Lei Li}, {and}
  \bibinfo{person}{Chunhua Shen}.} \bibinfo{year}{2020}\natexlab{a}.
\newblock \showarticletitle{Solov2: Dynamic and fast instance segmentation}.
\newblock \bibinfo{journal}{\emph{Advances in Neural information processing
  systems}}  \bibinfo{volume}{33} (\bibinfo{year}{2020}),
  \bibinfo{pages}{17721--17732}.
\newblock


\bibitem[\protect\citeauthoryear{Xue, Zhang, Nishino, and Dana}{Xue
  et~al\mbox{.}}{2022}]%
        {Xue2022}
\bibfield{author}{\bibinfo{person}{Jia Xue}, \bibinfo{person}{Hang Zhang},
  \bibinfo{person}{Ko Nishino}, {and} \bibinfo{person}{Kristin~J. Dana}.}
  \bibinfo{year}{2022}\natexlab{}.
\newblock \showarticletitle{Differential Viewpoints for Ground Terrain Material
  Recognition}.
\newblock \bibinfo{journal}{\emph{IEEE Transactions on Pattern Analysis and
  Machine Intelligence}} \bibinfo{volume}{44}, \bibinfo{number}{3}
  (\bibinfo{year}{2022}), \bibinfo{pages}{1205--1218}.
\newblock
\urldef\tempurl%
\url{https://doi.org/10.1109/TPAMI.2020.3025121}
\showDOI{\tempurl}


\bibitem[\protect\citeauthoryear{Yang, Yang, Fu, Lu, and Guo}{Yang
  et~al\mbox{.}}{2020}]%
        {yang2020learning}
\bibfield{author}{\bibinfo{person}{Fuzhi Yang}, \bibinfo{person}{Huan Yang},
  \bibinfo{person}{Jianlong Fu}, \bibinfo{person}{Hongtao Lu}, {and}
  \bibinfo{person}{Baining Guo}.} \bibinfo{year}{2020}\natexlab{}.
\newblock \showarticletitle{Learning texture transformer network for image
  super-resolution}. In \bibinfo{booktitle}{\emph{Proceedings of the IEEE/CVF
  conference on computer vision and pattern recognition}}.
  \bibinfo{pages}{5791--5800}.
\newblock


\bibitem[\protect\citeauthoryear{Yuan, Huang, Guo, Zhang, Chen, and Wang}{Yuan
  et~al\mbox{.}}{2018}]%
        {yuan2018ocnet}
\bibfield{author}{\bibinfo{person}{Yuhui Yuan}, \bibinfo{person}{Lang Huang},
  \bibinfo{person}{Jianyuan Guo}, \bibinfo{person}{Chao Zhang},
  \bibinfo{person}{Xilin Chen}, {and} \bibinfo{person}{Jingdong Wang}.}
  \bibinfo{year}{2018}\natexlab{}.
\newblock \showarticletitle{Ocnet: Object context network for scene parsing}.
\newblock \bibinfo{journal}{\emph{arXiv preprint arXiv:1809.00916}}
  (\bibinfo{year}{2018}).
\newblock


\bibitem[\protect\citeauthoryear{Zhang, Gu, Zhang, Bao, Chen, Wen, Wang, and
  Guo}{Zhang et~al\mbox{.}}{2022}]%
        {zhang2022styleswin}
\bibfield{author}{\bibinfo{person}{Bowen Zhang}, \bibinfo{person}{Shuyang Gu},
  \bibinfo{person}{Bo Zhang}, \bibinfo{person}{Jianmin Bao},
  \bibinfo{person}{Dong Chen}, \bibinfo{person}{Fang Wen},
  \bibinfo{person}{Yong Wang}, {and} \bibinfo{person}{Baining Guo}.}
  \bibinfo{year}{2022}\natexlab{}.
\newblock \showarticletitle{Styleswin: Transformer-based gan for
  high-resolution image generation}. In \bibinfo{booktitle}{\emph{Proceedings
  of the IEEE/CVF Conference on Computer Vision and Pattern Recognition}}.
  \bibinfo{pages}{11304--11314}.
\newblock


\bibitem[\protect\citeauthoryear{Zhang, Dauphin, and Ma}{Zhang
  et~al\mbox{.}}{2019}]%
        {zhang2019fixup}
\bibfield{author}{\bibinfo{person}{Hongyi Zhang}, \bibinfo{person}{Yann~N
  Dauphin}, {and} \bibinfo{person}{Tengyu Ma}.}
  \bibinfo{year}{2019}\natexlab{}.
\newblock \showarticletitle{Fixup initialization: Residual learning without
  normalization}.
\newblock \bibinfo{journal}{\emph{arXiv preprint arXiv:1901.09321}}
  (\bibinfo{year}{2019}).
\newblock


\bibitem[\protect\citeauthoryear{Zhou, Zhao, Puig, Fidler, Barriuso, and
  Torralba}{Zhou et~al\mbox{.}}{2017}]%
        {zhou2017scene}
\bibfield{author}{\bibinfo{person}{Bolei Zhou}, \bibinfo{person}{Hang Zhao},
  \bibinfo{person}{Xavier Puig}, \bibinfo{person}{Sanja Fidler},
  \bibinfo{person}{Adela Barriuso}, {and} \bibinfo{person}{Antonio Torralba}.}
  \bibinfo{year}{2017}\natexlab{}.
\newblock \showarticletitle{Scene parsing through ade20k dataset}. In
  \bibinfo{booktitle}{\emph{Proceedings of the IEEE conference on computer
  vision and pattern recognition}}. \bibinfo{pages}{633--641}.
\newblock


\bibitem[\protect\citeauthoryear{Zhou, Hasan, Deschaintre, Guerrero,
  Sunkavalli, and Kalantari}{Zhou et~al\mbox{.}}{2022}]%
        {zhou2022tilegen}
\bibfield{author}{\bibinfo{person}{Xilong Zhou}, \bibinfo{person}{Milos Hasan},
  \bibinfo{person}{Valentin Deschaintre}, \bibinfo{person}{Paul Guerrero},
  \bibinfo{person}{Kalyan Sunkavalli}, {and} \bibinfo{person}{Nima~Khademi
  Kalantari}.} \bibinfo{year}{2022}\natexlab{}.
\newblock \showarticletitle{TileGen: Tileable, Controllable Material Generation
  and Capture}. In \bibinfo{booktitle}{\emph{SIGGRAPH Asia 2022 Conference
  Papers}}. \bibinfo{pages}{1--9}.
\newblock


\end{thebibliography}

\end{document}